\pdfoutput=1
\documentclass[11pt, logo, a4paper, copyright, nonumbering]{nvidiatechreport}
\renewcommand{\today}{}  %

\usepackage{url}
\usepackage{booktabs}
\usepackage{amsfonts}
\usepackage{amssymb}
\usepackage{amsmath}
\usepackage{nicefrac}
\usepackage{microtype}
\usepackage{xcolor}
\usepackage{graphicx}
\usepackage{subcaption}
\usepackage{multirow}
\usepackage{wrapfig}
\usepackage{xspace}
\usepackage{enumitem}
\usepackage{array}
\usepackage{algorithm}
\usepackage{algpseudocode}
\usepackage{float}
\usepackage[authoryear]{natbib}
\usepackage{stfloats}
\usepackage[section]{placeins}
\usepackage{dirtree}

\usepackage{listingsutf8}
\usepackage[most]{tcolorbox}
\tcbuselibrary{listings,skins,breakable}
\newcommand{\name}[0]{\mbox{\textsc{ASPIRE}}\xspace}

\hypersetup{
  colorlinks=false,
  pdfborder={0 0 1}
}

\title{\MakeUppercase{\name}: Agentic \texttt{/Skills} Discovery for Robotics}

\author{
  Runyu Lu$^{1\;2\;*\;\dagger}$, Yubo Wu$^{1\;3\;*}$, Ethan Kou$^{1\;4\;*}$, Haotian Lin$^{1\;5\;*}$, 
  Letian Fu$^{1\;4}$, Wenli Xiao$^{1\;5}$, Ajay Mandlekar$^{1}$, Yinzhen Xu$^{1}$,
  Guanya Shi$^{5}$, Ken Goldberg$^{4}$, Ang Chen$^{2}$, Mosharaf Chowdhury$^{2}$,
  Jan Kautz$^{1}$, Yuke Zhu$^{1\;\dagger}$, Linxi ``Jim'' Fan$^{1\;\dagger}$, Guanzhi Wang$^{1\;\dagger}$ \\
  \small{$^1$ NVIDIA, $^2$ UMich, $^3$ UIUC, $^4$ UC Berkeley, $^5$ CMU} \\
  \small{${^*}$ Equal contribution, ${^\dagger}$ Project leads} \\
  \par\vspace{0.25em}
  \small\url{https://research.nvidia.com/labs/gear/aspire/}
}

\begin{abstract}
\textbf{Abstract:}

Traditional robot programming is notoriously challenging: it requires orchestrating multimodal perception, managing complex physical contact dynamics, and handling diverse environment configurations and execution failures.
We introduce \name (\textbf{\underline{A}}gentic \textbf{\underline{S}}kill \textbf{\underline{P}}rogramming through \textbf{\underline{I}}terative \textbf{\underline{R}}obot \textbf{\underline{E}}xploration), a continual learning system for robotics that autonomously writes and refines robot control programs in a code-as-policy paradigm while compounding experience into a reusable skill library. \name enables automated discovery of reusable skills that persist across multiple tasks, simulation and real-world settings, and different embodiments.
Rather than relying on fixed, human-engineered pipelines, \name operates in an open-ended learning loop, consisting of three key components: (1) a closed-loop robot execution engine that exposes fine-grained multimodal traces (e.g., perception overlays, grasp candidates, motion trajectories, and collision feedback), enabling the agent to autonomously diagnose failures, synthesize repairs, and validate outcomes; (2) a continually expanding skill library that distills validated fixes into reusable, transferable robotic knowledge; and (3) an evolutionary search procedure that generates diverse programs, systematically debugging them to explore beyond single-trajectory refinement. As \name encounters more tasks, its growing skill library enables increasingly rapid adaptation.
Consequently, \name surpasses prior methods by up to 77\% on manipulation tasks under perturbation, 72\% on Robosuite's bimanual handover task, and up to 32\% on long-horizon household tasks. The accumulated skill library further enables strong zero-shot generalization: on representative unseen long-horizon tasks, \name achieves 31\% success, substantially outperforming the 4\% success rate of prior methods despite their heavy reliance on test-time reasoning and retries. 
Performance scales with the capabilities of frontier models. With Claude Opus~5.5, \name\ nearly saturates these benchmarks, achieving 99\% average success across all 80 LIBERO-Pro tasks and 100\% success on representative Robosuite and BEHAVIOR-1K tasks. Beyond simulation, skills discovered in simulation reduce real-world robot programming effort across embodiments and robot APIs, while skills acquired through online interaction with physical robots enable \name\ to solve a Rubik's Cube with progressively improving efficiency.
\end{abstract}

\begin{document}

\maketitle

\section{Introduction}
\label{sec:intro}

\looseness=-1
Recent progress in software engineering agents demonstrates that language models can autonomously inspect execution traces, localize failures, revise implementations, and improve through repeated interaction with execution environments~\citep{claudecode2025, openai2025codex, opencode2025, wang2025openhands, yang2024sweagent}.
In robotics, this paradigm has inspired \emph{code-as-policy} systems that compose perception modules, planning APIs, and control primitives into executable robot programs~\citep{liang2023code, singh2023progprompt, ahn2022saycan, huang2023voxposer, mu2024robocodex, capx2026}.
Because robot behaviors are represented explicitly as programs, they can in principle be inspected, edited, debugged, and refined through interaction feedback.

However, existing robotic coding agents remain fundamentally limited by naive execution environments that provide only coarse task-level feedback.
Debugging robot programs is intrinsically challenging because failures can arise from many interacting components, including multimodal perception, motion planning, grasp generation, contact dynamics, and long-horizon task coordination.
A failed rollout may indicate that the task did not succeed, but not whether the root cause is incorrect perception, an unstable grasp, a planning error, or a downstream recovery failure.
Without fine-grained diagnostic traces, agents have limited ability to determine what evidence to inspect, how to localize failures, or what repair strategy to attempt.

Moreover, existing systems do not accumulate experience across tasks.
Once a task is completed, discovered fixes and recovery strategies are discarded rather than consolidated into reusable skills.
As a result, the agent solving its hundredth task is effectively no more experienced than the agent solving its first.

Human robotics engineers take a fundamentally different approach.
When a robot program fails, they replay executions, inspect perception outputs and motion trajectories, localize the failing subsystem, revise the implementation, and internalize reusable recovery strategies.
Over time, debugging experience compounds into transferable knowledge, including grasp recovery heuristics, navigation strategies, prompting recipes, and procedural fixes that generalize across tasks.
This accumulation of reusable knowledge is a key reason why human robot programmers become progressively more effective over time.

\begin{figure*}[t!]
  \centering
  \includegraphics[width=\textwidth]{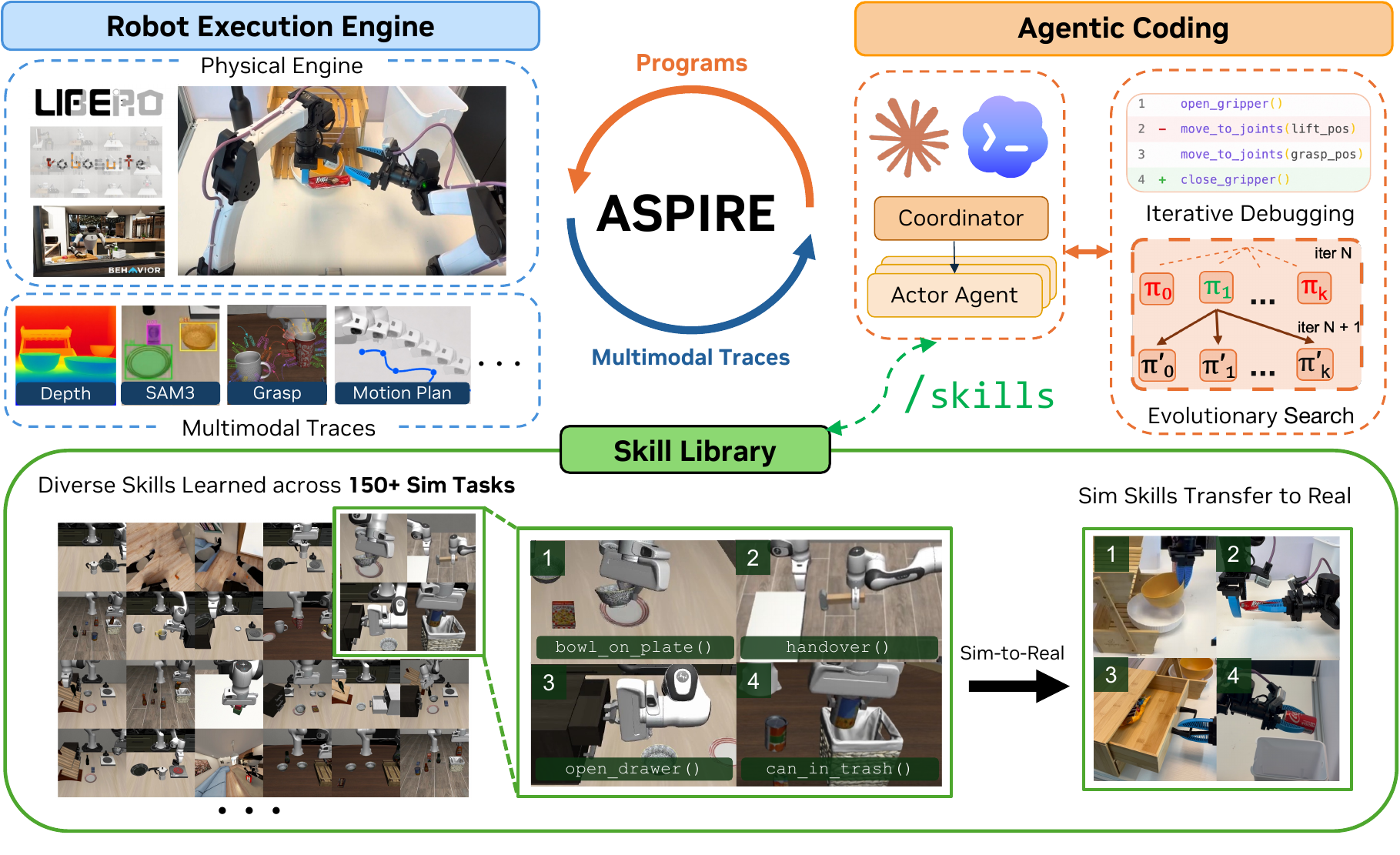}
  \caption{\textbf{\name system overview.}
  A \emph{coordinator} spawns an \emph{actor agent} (coding agent) per task, enabling parallel learning across tasks.
  Each actor refines and validates robot programs through \emph{iterative debugging} with the \emph{robot execution engine}, which exposes per-primitive multimodal traces for failure attribution and repair.
  \emph{Evolutionary search} samples diverse candidate programs ($\pi_0, \ldots, \pi_k$), sends each through the engine, and conditions the next generation ($\pi'_0, \ldots, \pi'_k$) on surviving programs and residual failure traces.
  The \emph{coordinator} writes validated repairs into a shared \emph{skill library}, which future actors retrieve as in-context guidance; skills discovered in simulation can also be adapted as cross-embodiment guidance for real-robot programming.
  }
  \label{fig:system-design}
\end{figure*}

In this work, we introduce \name, a self-improving continual learning robotic system that autonomously writes and refines robot control programs in a code-as-policy paradigm while accumulating experience into a reusable skill library.
Rather than operating within a fixed perception-plan-execute pipeline, \name runs in an open-ended learning loop, in which the agent self-determines how to inspect execution traces, diagnose failures, synthesize repairs, validate corrected behaviors, and consolidate successful recovery patterns into persistent skills that transfer across tasks.

An overview of \name is shown in Fig.~\ref{fig:system-design}.
\name is built from three components.
First, a \textbf{closed-loop robot execution engine} replaces coarse rollout-level feedback with per-primitive execution traces. For each perception, planning, grasping, and control call, the execution engine records the observations, inputs, outputs, and visual evidence if possible. These rich multimodal traces allow the agent to selectively inspect salient primitive logs, progressively localize failures, and validate repairs through re-execution.
Second, \name maintains a growing \textbf{skill library} that distills validated fixes into reusable, transferable robotic knowledge retrievable as in-context guidance for future tasks.
Third, \name employs an \textbf{evolutionary search procedure} that generates diverse programs, exploring beyond single-trajectory self-improvement through iterative debugging and parallel refinement.
Together, these components establish a self-improving robotic system whose performance scales with experience: the more tasks \name sees, the larger its skill library grows, and the more it transfers to novel tasks, longer-horizon behaviors, and real-world robotic settings where similar failure-recovery patterns emerge under different embodiments.

Empirically, \name\ improves success over prior coding agents~\citep{capx2026} by up to 77\%, and with frontier models it nearly saturates standard simulation benchmarks: with Claude Opus~5.5, it reaches 99\% average success across all 80 LIBERO-Pro~\citep{zhou2025liberopro} tasks and 100\% success on representative Robosuite~\citep{zhu2020robosuite} and BEHAVIOR-1K~\citep{behavior1k2023} tasks, and its success scales with the capability of the underlying frontier model.
A skill library accumulated on LIBERO-90~\citep{libero2024} transfers zero-shot to LIBERO-Pro Long (31\% versus 4\% for prior coding agents), and stronger models discover new strategies, such as knocking a book flat to push it under a shelf.
On a real bimanual robot, simulation-discovered skills reduce debugging effort, and skills learned online let \name\ solve a Rubik's Cube.

\section{Method}
\label{sec:method}

\begin{figure*}[t!]
  \centering
  \includegraphics[width=\textwidth]{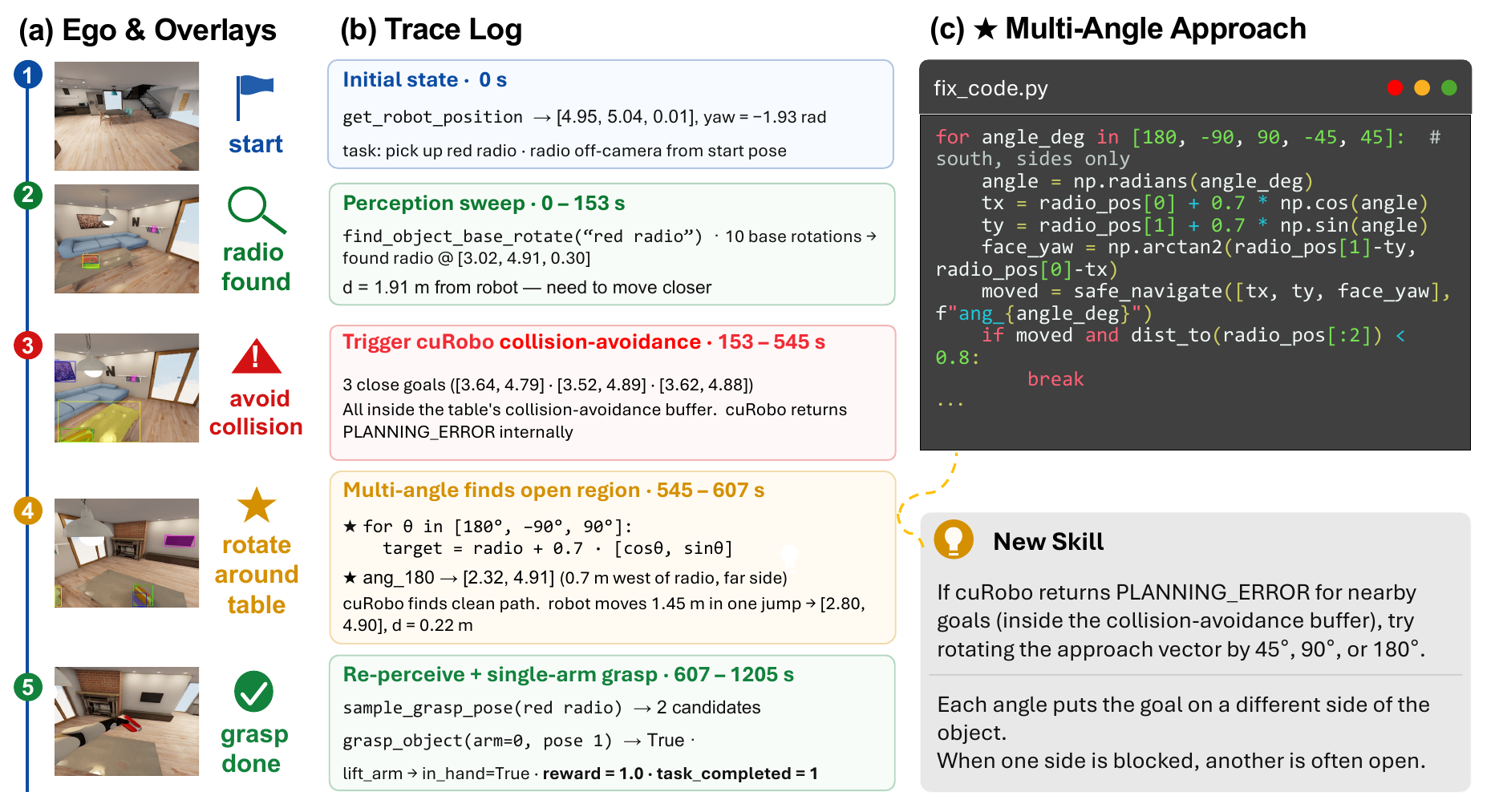}
  \caption{\textbf{Robot execution engine.}
  Trace-guided debugging on a BEHAVIOR-1K \textsc{navigate-and-pick-up-radio} task.
  (a)~Ego-view keyframes and overlays show the robot locating the radio but failing to approach it.
  (b)~The primitive trace localizes the failure to repeated \texttt{PLANNING\_ERROR}s: candidate navigation goals fall inside the table's collision-avoidance buffer.
  (c)~The agent patches the program with a multi-angle approach routine, re-perceives the radio from a reachable side, and completes the grasp. The validated repair is admitted as a reusable \emph{Multi-Angle Approach} skill.
  }
  \vspace{-0.45em}
  \label{fig:dojo}
\end{figure*}

\begin{figure*}[p]
  \centering
  \includegraphics[width=0.98\textwidth,height=0.76\textheight,keepaspectratio]{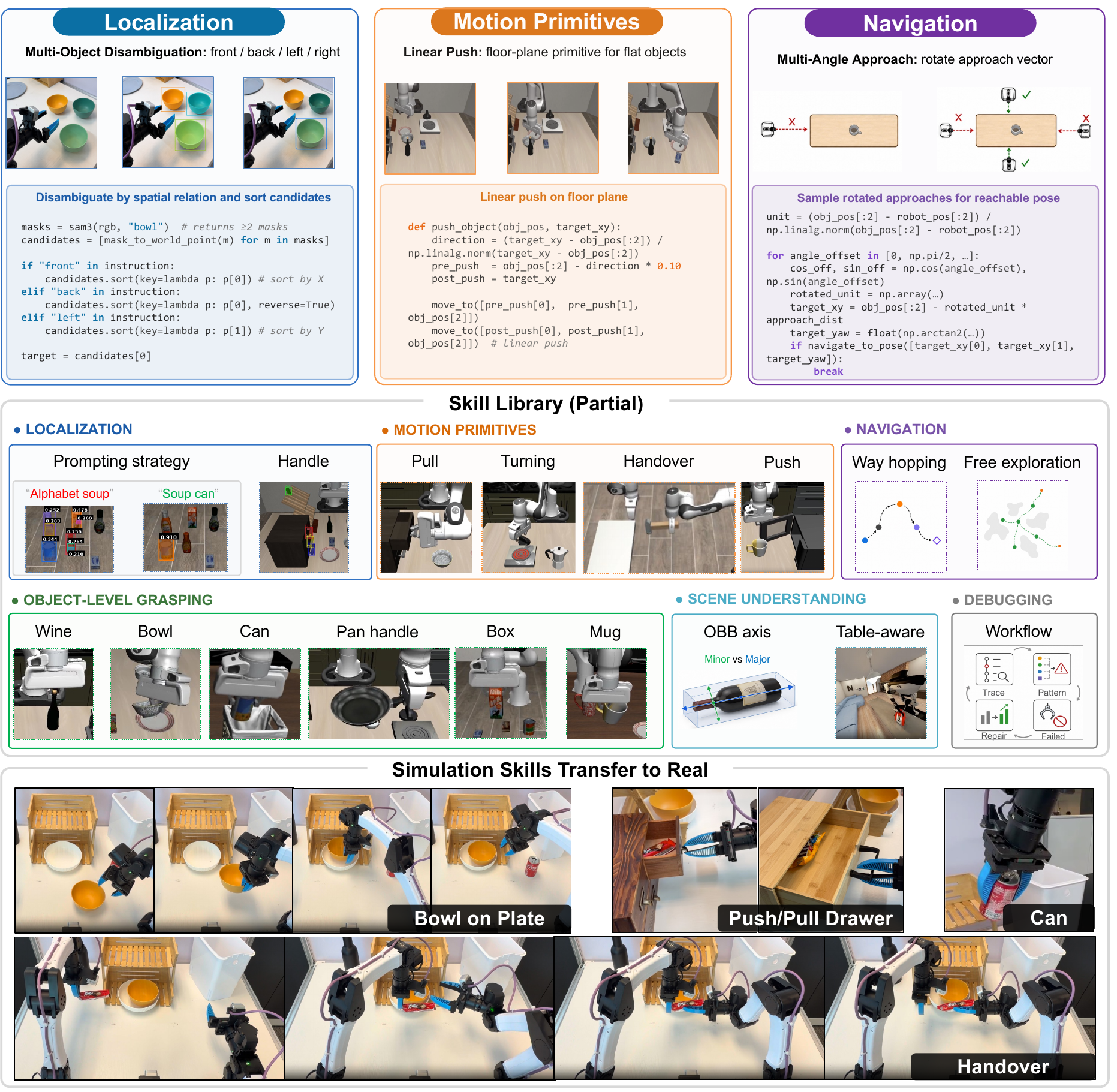}
  \caption{\textbf{Skill library.}
  \name stores validated, agent-discovered repair knowledge as reusable in-context skills rather than a fixed set of human-written primitives.
  Top: representative entries show learned skills about localization disambiguation, motion-primitive construction, and navigation recovery.
  Middle: the library grows across heterogeneous categories, including localization, navigation, motion primitives, object-level grasping, scene understanding, and debugging workflows.
  Bottom: selected skills discovered in simulation are used as in-context guidance for real-robot programming, providing evidence that skills can transfer across embodiments.
  Additional skill examples and prompts are shown in Appendix~\ref{sec:appendix:skill-library}.
  }
  \vspace{-0.45em}
  \label{fig:skill-library}
\end{figure*}

\name consists of three components that together form an open-ended learning loop: (1)~a \emph{robot execution engine} (§\ref{sec:method:dojo}) that exposes per-primitive multimodal traces for failure attribution and executes agent-written repairs for closed-loop validation; (2)~a \emph{skill library} (§\ref{sec:method:library}) that accumulates validated repairs as reusable skills; and (3)~an \emph{evolutionary search} procedure (§\ref{sec:method:evo}) that broadens exploration beyond single-trajectory repair. A coordinator also orders tasks as a curriculum and retries low-scoring tasks with the final library (§\ref{sec:method:verifier}). As \name encounters more tasks, its skill library grows, allowing future tasks to inherit accumulated repairs and reusable strategies.

As shown in Fig.~\ref{fig:system-design}, \name adopts a coordinator--actor architecture. A central coordinator manages the shared skill library and dispatches actors (coding agents) to individual tasks, while each actor writes, executes, diagnoses, and repairs robot programs within the robot execution engine. Actors do not exchange full chat histories or raw rollout trajectories. Instead, transferable experience is distilled into the skill library, allowing each actor’s context window to remain focused on the task specification, current program, and structured execution traces associated with the current failure.

\subsection{Robot Execution Engine}
\label{sec:method:dojo}

Embodied coding agents need execution evidence to debug robot programs. Prior methods expose this evidence through fixed, human-designed interfaces, typically as manually curated scene-level summaries or a set of pre-defined observations. This creates a trade-off: too little evidence can hide the failing primitive, while too much raw visual context can distract the agent from the causal chain that produced the failure.

\name turns this fixed feedback channel into an open-ended debugging environment. The robot execution engine records per-primitive multimodal traces for perception, planning, and control calls, exposes the trace to the actor, and executes actor-written repairs for closed-loop validation. We call this loop of execution, diagnosis, and repair \emph{iterative debugging}. For each primitive call, the trace stores the invoked API, inputs and outputs, return status, and relevant multimodal evidence such as RGB keyframes, overlays, grasp candidates, object poses, and motion-planning results. The actor does not receive full video frames; the engine keeps frames immediately before and after each primitive call together with the corresponding overlays and return values, so the actor can focus on evidence around calls implicated by the failure.

Fig.~\ref{fig:dojo} zooms in on one BEHAVIOR-1K \textsc{navigate-and-pick-up-radio} debugging episode. The ego-view keyframes show that the robot finds the radio, repeatedly fails to approach it, and then succeeds after changing its approach direction. The primitive trace localizes the failure: perception succeeds and returns a radio pose, but repeated \texttt{navigate\_to\_pose} calls return \texttt{PLANNING\_ERROR}. By checking the navigation return values and associated logs, the actor finds that the generated navigation target lies too close to the table boundary, within about 20~cm of the table edge. This triggers collision avoidance and causes the planner to fail. Thus, the failure is not due to detecting or grasping the radio, but rather to the infeasibility of the target pose under the table's collision constraints.

The repair follows directly from this diagnosis. Rather than changing the perception prompt or grasp primitive, the actor writes a multi-angle approach routine that samples alternative navigation targets around the radio and executes an approach direction that clears the collision buffer. The execution engine exposes the evidence, validates the patched program, and enables the actor to analyze the resulting logs, form a hypothesis, and make a targeted repair decision.

\subsection{Skill Library}
\label{sec:method:library}

Program failures recur across tasks, but the reusable knowledge is rarely an entire task program. \name's skill library stores heterogeneous repair knowledge: localization heuristics, perception prompts, grasping constraints, navigation recovery strategies, motion primitives, scene-understanding routines, and debugging workflows. We do not prescribe this taxonomy in advance. Skills are induced from validated repairs: the actor diagnoses a failure from execution traces, patches the program, and validates the fix on the debug seeds.

Each skill is stored as compact in-context guidance, including the failure signature, when-to-apply condition, repair strategy, and, when useful, a representative code sketch. Fig.~\ref{fig:skill-library} shows representative entries and the resulting library breadth. For the radio task in Fig.~\ref{fig:dojo}, the admitted skill is a navigation recovery pattern rather than a complete radio-pickup program: when planner errors recur near an obstacle boundary because sampled target poses fall inside the collision buffer, sample alternative approach directions around the object before retrying perception and grasping. This representation lets future actors reuse validated repairs instead of rediscovering them through test-time reasoning, supports zero-shot transfer to harder simulated tasks, and provides the mechanism for selected simulation-discovered skills to generalize across embodiments and transfer to real robots.

Skill admission has two checks. Each actor reports structured findings: the failure mode, the validated fix, and any potentially transferable repair pattern as a proposed skill. A fresh verifier actor then receives the proposed skill and attempts the source task independently on a subset of seeds, recording which entries are sound, defective, or untested. The coordinator verifies compliance with the allowed API policy and promotes only reusable entries whose success the verifier reproduced. Appendix~\ref{sec:appendix:skill-library-prompts} provides the prompts for actor reporting and skill admission.

\subsection{Evolutionary Search}
\label{sec:method:evo}

Iterative debugging alone can collapse into local repair loops, where the actor repeatedly patches the same failed strategy instead of exploring fundamentally different ways to solve the task.
\name uses evolutionary search to broaden exploration of executable robot programs, encouraging diverse repair hypotheses and task strategies.

In each round, based on the skill library, the actor proposes a population of $K$ candidate programs conditioned on the top-performing previous programs and failure traces from previous evaluations.
Each candidate is executed in the robot execution engine, producing task outcomes together with new diagnostic traces.
The next round is then conditioned on the best-performing programs together with their remaining failure modes, allowing the search to explore distinct strategies rather than repeatedly refining the same solution.

The search target is the robot program itself.
Candidates are selected through closed-loop execution on the debug seeds, and validated repairs are admitted into the skill library after search concludes, provided they generalize across environment variations and tasks. Search terminates when the best candidate reaches a success threshold $\theta$ on the debug seeds or when the budget of $T$ rounds is exhausted; skill patterns are extracted from the debugging history, and the final program is then evaluated once on held-out evaluation seeds.
Algorithm~\ref{alg:evolutionary-repair} summarizes the evolutionary search procedure. See Appendix~\ref{sec:appendix:prompts-skills} for the pipeline files.

\subsection{Curriculum and Backward Transfer}
\label{sec:method:verifier}

The coordinator also acts as a curriculum agent. Based on the current skill library, it chooses the next task to attempt from the list. It can attempt to fill a library gap, give an existing skill a chance to work on a new task, or utilize a recent finding by picking tasks in the same suite. 
Because tasks are attempted sequentially, those attempted early have access to only a small library, while later tasks benefit from every skill learned before them. An early low score may reflect missing skills rather than task difficulty. Therefore, once all tasks are finished and the skill library is complete, the coordinator allows one separate rerun of low-scoring tasks that used an earlier version of the library. This backward skill transfer allows skills to be applied to tasks that came before them.

\begin{algorithm}[t]
\scriptsize
\caption{Evolutionary search over programs}
\label{alg:evolutionary-repair}
\begin{algorithmic}[1]
\Require task $\tau$, program $P^0$, debug/held-out seeds $S_{\rm dbg},S_{\rm eval}$, skill library $\mathcal L$, agent $M$, budget $(K,T)$, threshold $\theta$
\Statex \textbf{Notation:} $\textsc{Execute}(P,S)=(r,Z)$ returns score $r$ and trace bundle $Z$.
\State $(r^\star,Z^0)\gets\textsc{Execute}(P^0,S_{\rm dbg})$; \quad $P^\star\gets P^0$
\State $\mathcal H\gets\{(P^0,r^\star,Z^0)\}$
\For{$i=1,\ldots,T$}
    \State $\{P_i^k\}_{k=1}^K\gets{}$
    \Statex \hspace{\algorithmicindent}$\textsc{ProposeRepairs}(M,\tau,\mathrm{Top3}(\mathcal H),\mathcal L,\mathcal H)$
    \For{$k=1,\ldots,K$} \State $(r_i^k,Z_i^k)\gets\textsc{Execute}(P_i^k,S_{\rm dbg})$
    \EndFor
    \State $\mathcal H\gets\mathcal H\cup\{(P_i^k,r_i^k,Z_i^k)\}_{k=1}^K$; \quad $k^\star\gets\arg\max_k r_i^k$
    \If{$r_i^{k^\star}>r^\star$}
        \State $(P^\star,r^\star)\gets(P_i^{k^\star},r_i^{k^\star})$
    \EndIf
    \If{$r^\star\ge\theta$} \State \textbf{break} \EndIf
\EndFor
\State $\mathcal G\gets\textsc{ExtractValidatedPatterns}(\mathcal H,P^\star)$ \Comment{debug history only}
\State $(r_{\rm eval},\cdot)\gets\textsc{Execute}(P^\star,S_{\rm eval})$ \Comment{run once, after search}
\State \Return $(P^\star,r_{\rm eval},\mathcal G)$
\end{algorithmic}
\end{algorithm}

\section{Experiments}
\label{sec:experiments}

\subsection{Experimental Setup}
\label{sec:experiments:setup}
For the main results, we use Claude Code~\citep{claudecode2025} with Claude Opus~4.6~\citep{anthropic2026opus46} and a 1M-token context window as the coding agent on all simulation benchmarks. The agent writes executable Python robot programs in \name's unified code-as-policy harness, which exposes the same robot programming APIs for perception, geometry, and motion planning across LIBERO-Pro, Robosuite, and BEHAVIOR-1K. The environment and API set are fixed across all simulation experiments. The model ablation (\S\ref{sec:experiments:ablation}) compares twelve models: Claude Opus~4.6, Opus~5~\citep{anthropic2026opus5}, Opus~5.5~\citep{anthropic2026opus55}, Sonnet~5.5~\citep{anthropic2026sonnet55}, Haiku~5.5~\citep{anthropic2026haiku55}, and Fable~5.1~\citep{anthropic2026fable51}; \mbox{GPT-6} Astra~\citep{openai2026gpt6astra}, \mbox{GPT-6} Sol and \mbox{GPT-6} Luna~\citep{openai2026gpt6sol}, and \mbox{GPT-6.1} Sol~\citep{openai2026gpt61sol}; and the open-weight Kimi~K3~\citep{kimiteam2026kimik3} and Qwen~3.8 Max~\citep{qwen2026qwen38max}. Claude and open-weight models run through Claude Code and GPT models through Codex CLI~\citep{openai2025codex}. The case study (\S\ref{sec:experiments:case-study}) uses Claude Opus~4.6, Opus~5, Opus~5.5, and \mbox{GPT-6} Astra.

For the real-robot skill transfer study (\S\ref{sec:experiments:real-robot}), we use OpenAI Codex~\citep{openai2025codex} with \mbox{GPT-5.5}~\citep{openai2026gpt55} in reasoning-xhigh mode on a bimanual YAM manipulation station. We select three skills compiled by \name in Franka-based simulation: soda-can pickup, bowl-on-plate placement, and drawer push/pull. We provide these as in-context guidance to the real-robot coding agent. The real-robot setting uses a different embodiment and API from simulation, but exposes \name multimodal execution traces so the agent can autonomously run, inspect, and debug programs without task-specific human guidance during iterative debugging. We compare runs with and without the corresponding simulation-discovered skill, measuring tokens to first success and held-out real-robot success rate. For online skill learning on the real robot (\S\ref{sec:cube-agentic-skill-learning}), \mbox{GPT-6} Astra in Codex solves a Rubik's Cube on the same bimanual YAM station.

\begin{figure*}[t!]
    \centering
    \includegraphics[width=0.85\textwidth]{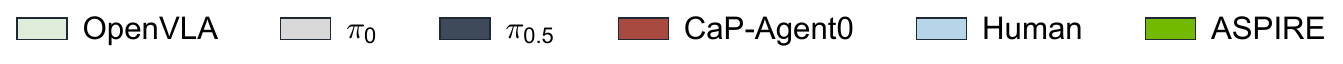}

    \vspace{-0.2em}

    \begin{subfigure}{\textwidth}
        \centering
        \includegraphics[width=\linewidth]{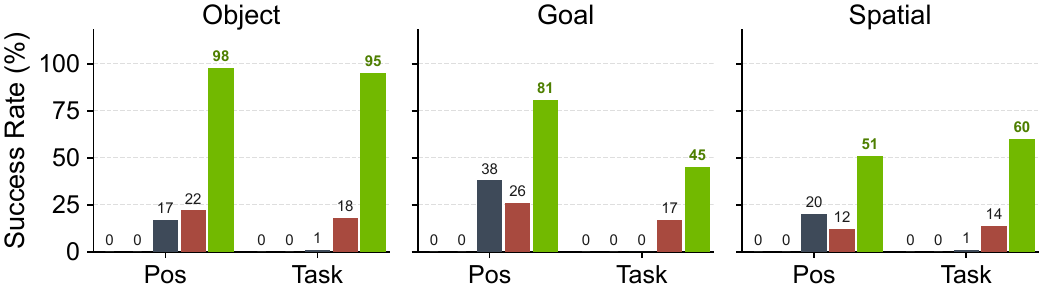}
        \captionsetup{justification=raggedright,singlelinecheck=false}
        \caption{\textbf{LIBERO-Pro}~\citep{zhou2025liberopro}: macro-averaged success over 10 tasks $\times$ 50 held-out environment seeds per suite/perturbation. For each task, \name\ learns and collects skills on seeds 51--65 and evaluates one generated program on seeds 1--50.}
        \label{fig:main-libero-pro}
        \captionsetup{justification=centering,singlelinecheck=true}
   
    \end{subfigure}

    \vspace{0.2em}

    \begin{subfigure}[t]{0.48\textwidth}
        \centering
        \includegraphics[width=\linewidth]{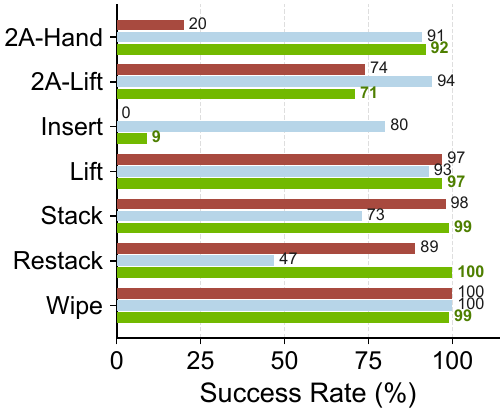}
        \caption{\textbf{Robosuite}~\citep{zhu2020robosuite}: success over 100 held-out trials per task. \name\ learns on seeds 101--125 and evaluates one generated program per task on seeds 1--100; 2A tasks are bimanual.}
        \label{fig:main-robosuite}
    \end{subfigure}%
    \hfill
    \begin{subfigure}[t]{0.48\textwidth}
        \centering
        \includegraphics[width=\linewidth]{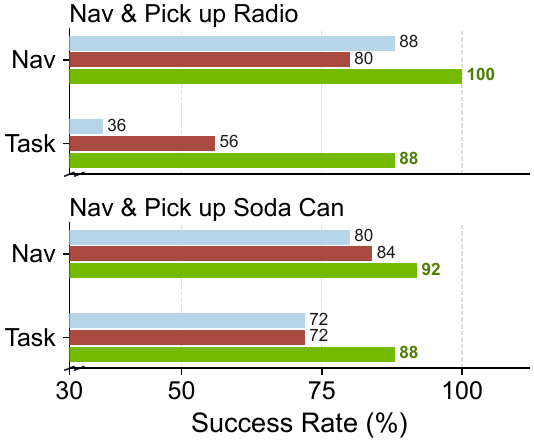}
        \caption{\textbf{BEHAVIOR-1K}~\citep{behavior1k2023}: long-horizon mobile manipulation on two household tasks. \name\ learns on seeds 26--35 and evaluates on seeds 1--25 with incremental block execution; Navigation and Task success are reported separately.}
        \label{fig:main-b1k}
    \end{subfigure}

    \caption{\textbf{\name\ improves over prior coding agents and end-to-end VLAs across three benchmark families.} (a)~Short-horizon manipulation on LIBERO-Pro; (b)~contact-rich manipulation on Robosuite; (c)~long-horizon mobile manipulation on BEHAVIOR-1K. \name\ evaluates one generated program per task across held-out seeds, while CaP-Agent0 regenerates a separate program per seed with test-time reasoning and retries. \name improves performance across all benchmarks, with several results surpassing programs written by human experts. Detailed per-task results are in Appendices~\ref{sec:appendix:main-results-tables} and~\ref{sec:appendix:ablations:swap}.
    }
    \vspace{-0.45em}
    \label{fig:main-results}
\end{figure*}

\subsection{Benchmarks and Baselines}
\label{sec:experiments:benchmarks}
We evaluate \name\ on three benchmark families: LIBERO-Pro~\citep{zhou2025liberopro} for short-horizon robustness under object, goal, and spatial perturbations; Robosuite~\citep{zhu2020robosuite} for contact-rich single- and dual-arm manipulation; and BEHAVIOR-1K~\citep{behavior1k2023} for long-horizon household mobile manipulation on \textsc{navigate-and-pick-up-soda-can} and \textsc{navigate-and-pick-up-radio}.
Our primary coding-agent baseline is CaP-Agent0~\citep{capx2026}, which uses visual differencing, a predefined skill library, and per-episode test-time retries.
We also compare with end-to-end VLA policies, including OpenVLA~\citep{kim2024openvlaopensourcevisionlanguageactionmodel}, $\pi_0$~\citep{black2024pi0visionlanguageactionflowmodel}, and $\pi_{0.5}$~\citep{intelligence2025pi05}.
For zero-shot transfer, we evaluate held-out LIBERO-Pro Long tasks using the skill library accumulated on LIBERO-90~\citep{libero2024}.

\subsection{Evaluation Protocol}
Across all benchmarks, an environment seed fixes each task instance, including object poses, distractors, and initial robot/object states. We use disjoint debug and evaluation seeds: \name\ learns on a small debug split, then reports success on larger held-out evaluation seeds with one generated program per LIBERO-Pro/Robosuite task, while CaP-Agent0 regenerates a separate program for each seed with test-time reasoning and retries. For BEHAVIOR-1K evaluation, \name\ uses incremental block execution, generating each next code block from the current multimodal trace.

\subsection{Main Evaluation Results}
\label{sec:experiments:main}

Figure~\ref{fig:main-results} summarizes the main evaluation. ``Human'' means the programs are written by human experts. On LIBERO-Pro, \name\ improves over all the baselines on all three suites: averaging position and task perturbations, \name\ gains 77\% on Object, 41.5\% on Goal, and 42.5\% on Spatial over the strongest baseline in each suite. $\pi_{0.5}$ is stronger than OpenVLA and $\pi_0$ on some position perturbations, but remains far below \name\ and largely collapses under task paraphrases. On Robosuite, \name\ preserves near-saturated performance on easier contact-rich tasks and substantially improves bimanual handover, from 20\% to 92\%. On BEHAVIOR-1K, \name\ outperforms both human and CaP-Agent0 on navigation and task success, with the largest task-level gain over CaP-Agent0 on \textsc{navigate-and-pick-up-radio}, from 56\% to 88\%.

\subsection{Zero-Shot Transfer to Unseen Tasks}
\label{sec:experiments:transfer}

We evaluate whether repair skills accumulated on LIBERO-90 transfer zero-shot to held-out LIBERO-Pro Long tasks. \name builds library snapshots from $N \in \{0,25,50,90\}$ source tasks, where $N{=}0$ is an empty-library setting and $N{=}90$ is the full suite. For each held-out task, \name\ generates one program and evaluates it across seeds with no additional debugging, retries, or task-specific library updates.

\begin{figure*}[t!]
\centering
\begin{subfigure}[t]{0.48\textwidth}
    \centering
    \includegraphics[width=\linewidth]{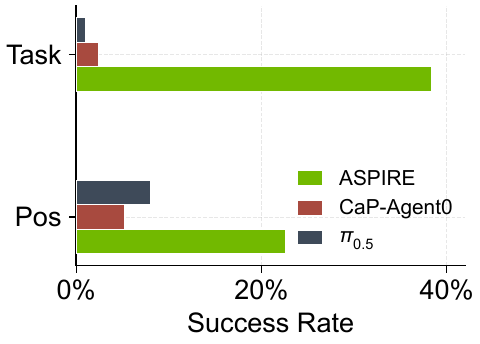}
    \caption{Full $N{=}90$ library zero-shot results versus baselines.}
\end{subfigure}%
\hfill
\begin{subfigure}[t]{0.48\textwidth}
    \centering
    \includegraphics[width=\linewidth]{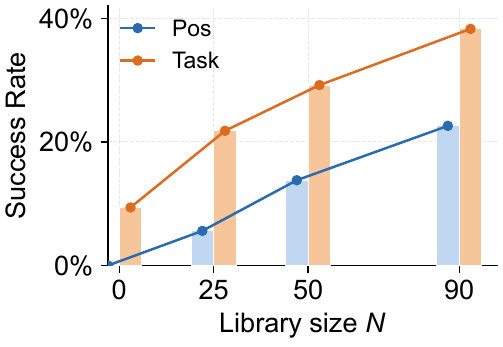}
    \caption{Zero-shot success improves as skill-library size increases.}
\end{subfigure}
\caption{\textbf{Cross-task zero-shot transfer on LIBERO-Pro Long.} Skills accumulated on LIBERO-90 improve zero-shot performance on held-out long-horizon tasks. Figure (a) compares the full $N{=}90$ library with baselines. Figure (b) shows success under position and task perturbations as the size of the skill library increases. All success rates are macro-averaged over 10 tasks per axis. Per-task results are in Appendix~\ref{sec:appendix:libero-long-pro-zero-shot}.}
\vspace{-0.45em}
\label{fig:transfer}
\end{figure*}

Figure~\ref{fig:transfer}(a) compares the full $N{=}90$ skill library with prior baselines on the two LIBERO-Pro Long transfer axes: \name reaches 23\% success on position perturbations and 38\% on task perturbations, outperforming CaP-Agent0 and $\pi_{0.5}$ on both axes. Figure~\ref{fig:transfer}(b) shows that success increases consistently as the size of the skill library grows, indicating that validated repairs from short-horizon tasks provide reusable robotic knowledge for longer-horizon compositions.

\subsection{Real-Robot Skill Transfer Across Embodiments}
\label{sec:experiments:real-robot}
We evaluate whether skills learned in simulation can reduce debugging effort on a real robot with a different embodiment. The transfer is not a direct policy deployment: the real robot uses its own perception, calibration, and control stack, and the coding agent must still adapt programs through real-world execution feedback. The question is whether retrieving simulation-discovered skills provides useful in-context guidance that reduces the amount of real-world debugging needed to reach a successful program.

Table~\ref{tab:real-robot-sim-to-real} shows that transferred skills consistently reduce debugging cost, while their effect on final success is task-dependent. Bowl placement succeeds in both settings but uses fewer tokens with skill retrieval; soda-can lifting improves from 65\% to 95\% while reducing total tokens by nearly an order of magnitude; and drawer manipulation reaches 55\% success with skill guidance, while the no-skill baseline exhausts a larger token budget without producing a successful evaluation program. These results indicate that selected failure-derived skills can guide real-robot program synthesis across embodiment and API changes rather than merely memorizing simulator-specific code.

\begin{table*}[t]
\centering
\caption{\textbf{Real-robot cross-embodiment skill transfer.} For each task, we compare real-robot debugging with and without retrieving a corresponding simulation-discovered skill from \name's library. Token counts are measured until the first successful real-robot program, or until the debugging budget is exhausted for runs that do not reach success. Success rate reports evaluation trials for the generated programs.
}
\label{tab:real-robot-sim-to-real}
\vspace{0.25em}
\begin{tabular}{lcccccc}
\toprule
\multirow{2}{*}{Task} & \multicolumn{2}{c}{Output Tokens (M) $\downarrow$} & \multicolumn{2}{c}{Total Tokens (M) $\downarrow$} & \multicolumn{2}{c}{Success Rate $\uparrow$} \\
\cmidrule(lr){2-3} \cmidrule(lr){4-5} \cmidrule(lr){6-7}
 & w/o Skills & w/ Skills & w/o Skills & w/ Skills & w/o Skills & w/ Skills \\
\midrule
Put bowl on plate & 0.05 & 0.04 & 8.65 & 5.11 & 20/20 & 20/20 \\
Lift soda can & 0.18 & 0.03 & 61.94 & 6.58 & 13/20 & 19/20 \\
Open/push drawer & 1.33 & 0.36 & 334.92 & 81.67 & 0/20 & 11/20 \\
\bottomrule
\end{tabular}
\vspace{-0.45em}
\end{table*}

\subsection{Online Skill Learning on a Real Robot}
\label{sec:cube-agentic-skill-learning}

Online adaptation is an important step toward robustness in deployment. Given a fixed set of tools, \name\ runs on a real robot to self-improve its tool-use strategy: the coding agent acts as a \texttt{system-2} that calls executable perception and action tools and records what it learns as text skills. We show this on \texttt{Solve-Rubik's-Cube} over three sequential real-world episodes in one GPT-6 Astra session, on a bimanual YAM platform with soft grippers (Appendix~\ref{app:aspire-system2-architecture}). Initially, the agent is given a randomly scrambled cube and human-provided task guidance. The agent needs to identify the color formation of all faces purely from visual observation and may call a symbolic solver to provide a move sequence, then it must realize the solution and verify the physical manipulations. During the first few attempts, the agent explores strategies for a successful face turn and summarizes its experience as text-based skills.
The accumulated skills cover two-layer support, single-layer isolation, face-axis alignment, collision-avoiding handovers, and state-aware slip recovery---learning through context refinement rather than model-weight updates.

\begin{figure}[tbp]
\centering
\begin{subfigure}[b]{0.54\textwidth}
    \centering
    \includegraphics[width=\linewidth]{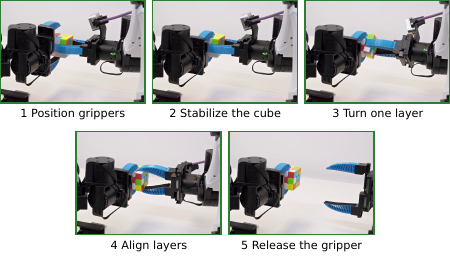}
    \caption{One face turn.}
    \label{fig:cube-face-turn}
\end{subfigure}\hfill
\begin{subfigure}[b]{0.43\textwidth}
    \centering
    \includegraphics[width=\linewidth]{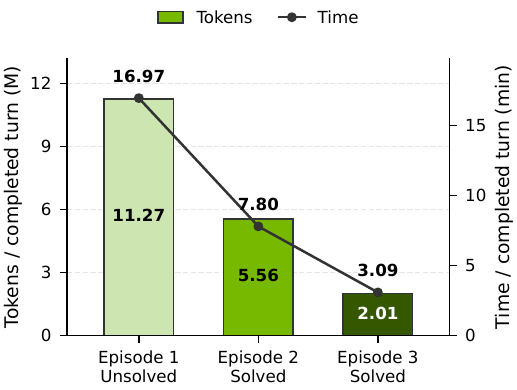}
    \caption{Efficiency across episodes.}
    \label{fig:cube-efficiency}
\end{subfigure}
\caption{\textbf{Solving a Rubik's Cube on the real robot.} (a)~GPT-6 Astra positions both grippers, holds the cube, turns one layer, aligns the layers, and releases. (b)~Time and tokens per completed turn fall across episodes as context and skills accumulate.}
\label{fig:cube-skill-efficiency}
\end{figure}

The first attempt remains unsolved after 50\,min\,54\,s; subsequent episodes solve the cube in 39\,min and 77\,min\,10\,s. Their logs report 3, 5, and 25 completed face turns, respectively, counting a half-turn as one move and excluding whole-cube reorientation. Figure~\ref{fig:cube-skill-efficiency}(a) shows one face turn, and Figure~\ref{fig:cube-skill-efficiency}(b) shows end-to-end time per completed turn decreasing from 16\,min\,58\,s to 7\,min\,48\,s and 3\,min\,05\,s. Across the first and final episodes, tokens per turn fall from 11.27\,M to 2.01\,M, and estimated model cost, at OpenAI API prices~\citep{openai2026pricing}, from \$16.85 to \$2.89. These full-episode amortized metrics include failed attempts, recovery, and context compaction. The results demonstrate increasingly efficient execution alongside accumulated skills.

\subsection{Ablation Studies}
\label{sec:experiments:ablation}

\begin{figure*}[t!]
    \centering
    \begin{subfigure}[t]{0.31\linewidth}
        \centering
        \includegraphics[width=\linewidth]{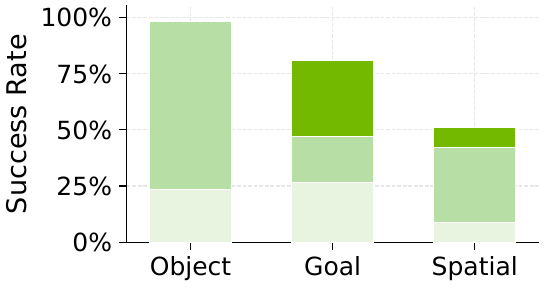}
        \caption{Position perturbation.}
    \end{subfigure}%
    \hfill
    \begin{subfigure}[t]{0.31\linewidth}
        \centering
        \includegraphics[width=\linewidth]{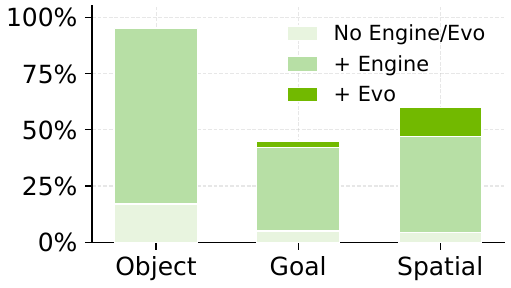}
        \caption{Task perturbation.}
    \end{subfigure}%
    \hfill
    \begin{subfigure}[t]{0.31\linewidth}
        \centering
        \includegraphics[width=\linewidth]{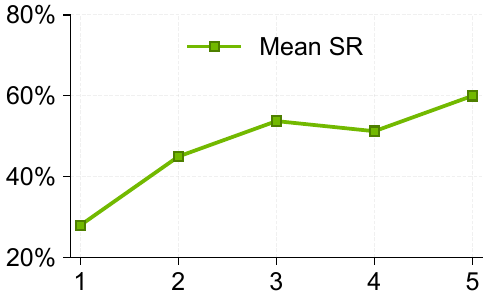}
        \caption{Per-iteration progress.}
    \end{subfigure}
    \caption{\textbf{Robot execution engine and evolutionary search ablations on LIBERO-Pro.} (a),~(b)~Stacked bars under position and task perturbation: the base system without the robot execution engine or evolutionary search, the gain from adding the engine, and the additional gain from evolutionary search. (c)~Average progress over low-performing tasks across evolutionary-search iterations. Per-task results are in Appendix~\ref{sec:appendix:ablations}.}
    \vspace{-0.45em}
    \label{fig:ablation-evo-dojo}
\end{figure*}

\paragraph{Robot execution engine and evolutionary search.}
We ablate two key components of \name\ on LIBERO-Pro: the robot execution engine, which exposes dense execution traces and validates task-level repairs, and evolutionary search, which explores additional repair candidates after iterative debugging with the engine. Figure~\ref{fig:ablation-evo-dojo}(a) and~(b) decompose success by component. The robot execution engine provides the largest average improvement, raising macro-average success from 14\% to 62\%; evolutionary search further improves the remaining hard tasks, reaching 72\% with both components.

\paragraph{Evolutionary search iterations.}
Figure~\ref{fig:ablation-evo-dojo}(c) tracks performance on low-performing tasks as the evolutionary-search budget increases. Success improves steadily in the first few rounds, suggesting that sampling multiple repair hypotheses quickly recovers alternatives missed by iterative debugging. The curve continues to increase more gradually afterward, indicating that additional rounds still help on residual hard cases, but with diminishing returns.

\paragraph{Backward Skill Transfer.}
We compare the initial and final held-out results of the five tasks selected for reruns with the full skill library in the Claude Opus~5 LIBERO-Pro Long campaign. Both evaluations use the same 50 seeded held-out layouts per task; the post-rerun column evaluates revised task programs after further debugging with the complete skill library.

\begin{table}[H]
\centering
\footnotesize
\caption{\textbf{Backward skill transfer on LIBERO-Pro Long.} Held-out Claude Opus~5 success before and after reruns. Each task has 50 held-out seeds. The five rerun tasks were selected for low initial success; task IDs give the perturbation (\texttt{pos}: position, \texttt{task}: task) and task index.}
\label{tab:libero10-verifier-rerun-ablation}
\setlength{\tabcolsep}{5pt}
\begin{tabular}{lccc}
\toprule
\textbf{Task} & \textbf{Initial} & \textbf{Post-rerun} & \textbf{$\Delta$ (pp)} \\
\midrule
\texttt{pos/0} Soup + sauce $\to$ basket & 37/50 (74\%) & 47/50 (94\%) & +20 \\
\texttt{pos/3} Bowl $\to$ drawer, close & 0/50 (0\%) & 26/50 (52\%) & +52 \\
\texttt{pos/9} Mug $\to$ microwave, close & 0/50 (0\%) & 0/50 (0\%) & 0 \\
\texttt{task/3} Bottle $\to$ drawer, close & 0/50 (0\%) & 36/50 (72\%) & +72 \\
\texttt{task/6} Red mug $\to$ plate, pudding right & 39/50 (78\%) & 43/50 (86\%) & +8 \\
\midrule
\textbf{All 20 tasks} & \textbf{746/1000 (74.6\%)} & \textbf{822/1000 (82.2\%)} & \textbf{+7.6} \\
\bottomrule
\end{tabular}
\end{table}

Four of the five selected tasks improve, raising suite-wide success from 74.6\% to 82.2\% (+7.6pp). The largest gains occur on the two drawer tasks, while \texttt{pos/9} remains at 0\%. The complete skill library lets the agent unlock success on two tasks and improve reliability on two more.

\paragraph{Model ablation.}
We vary only the LLM that drives the coding agent, keeping the environment, robot programming API, seed splits, and execution protocol fixed.

\begin{figure}[tbp]
\centering
\includegraphics[width=0.85\textwidth]{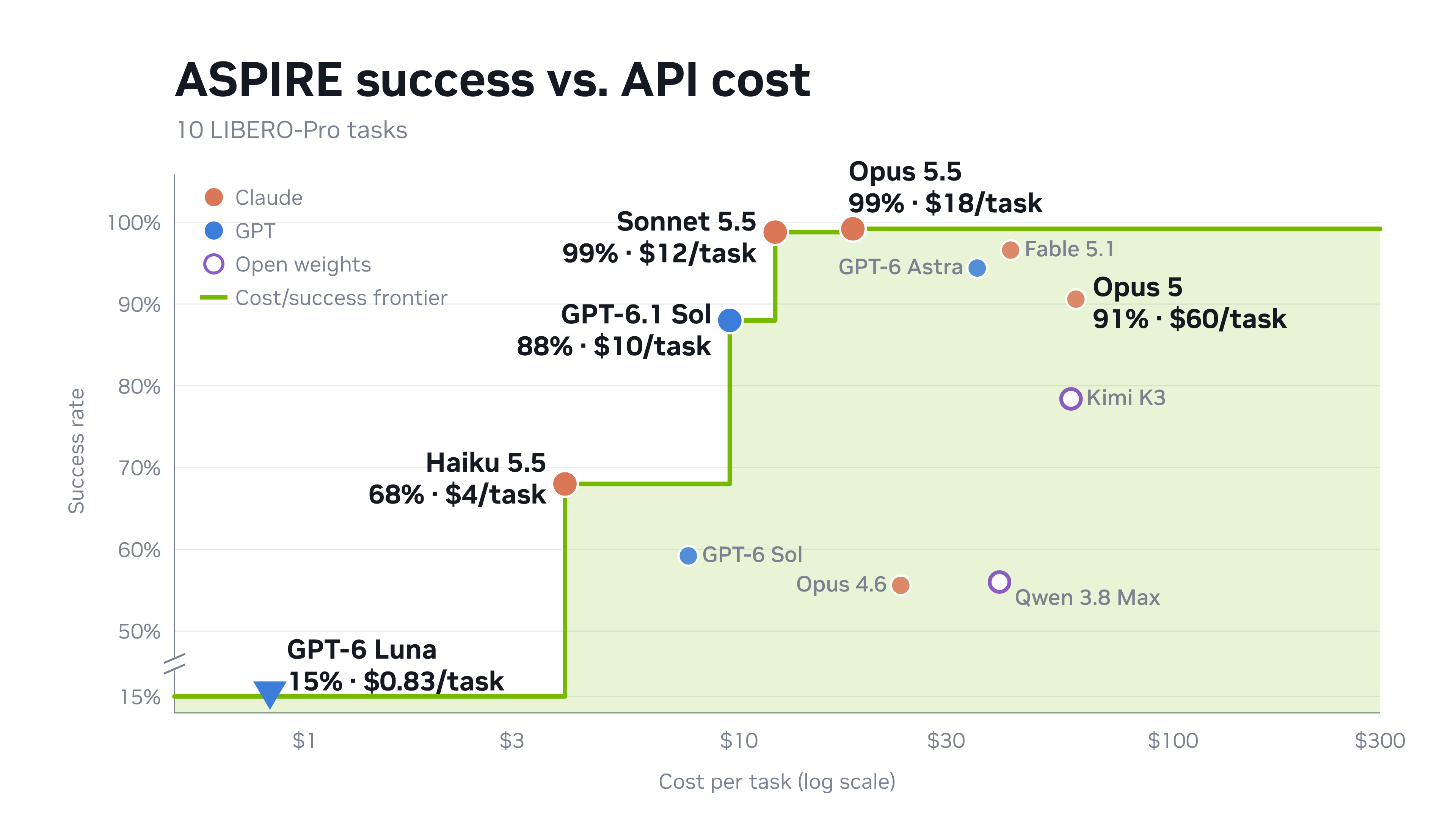}
\caption{\textbf{Model ablation: success versus cost.} Held-out success versus estimated API cost per task (list prices, log scale) on the 10 LIBERO-Pro Goal tasks under the position perturbation, using \name\ iterative debugging without evolutionary search and with an initially empty skill library. Colors mark the model family; hollow markers are the open-weight Kimi~K3 and Qwen~3.8 Max. The green line is the cost/success frontier: the models that no cheaper model outperforms, labeled with success and cost per task. The success axis is broken below 50\% to show GPT-6 Luna (15\%). All values are in Table~\ref{tab:model-cost-summary}; per-task results for Claude Opus~5, Opus~5.5, and GPT-6 Astra are in Appendix~\ref{sec:appendix:goal-swap-iterative-debug}.}
\label{fig:model-ablation}
\end{figure}

\smallskip\noindent\textbf{\name\ improves with the model.}\hspace{0.5em}On LIBERO-Pro Goal under the position perturbation (Figure~\ref{fig:model-ablation}), success rises from 15\% with GPT-6 Luna and 56\% with Claude Opus~4.6 to 99\% with Claude Opus~5.5 and Sonnet~5.5; the best open-weight model, Kimi~K3, reaches 78\%.

\smallskip\noindent\textbf{Cost/success Pareto frontier.}\hspace{0.5em}Stronger models are not necessarily more expensive. Claude Sonnet~5.5 is Pareto-optimal, reaching 98.8\% at \$12 per task, against \$18 for Opus~5.5 (99.2\%) and \$60 for Opus~5 (91\%).

\begin{figure}[tbp]
\centering
\captionsetup{font=small}
\begin{minipage}[t]{0.53\textwidth}
\vspace{0pt}
\centering
\includegraphics[width=\linewidth]{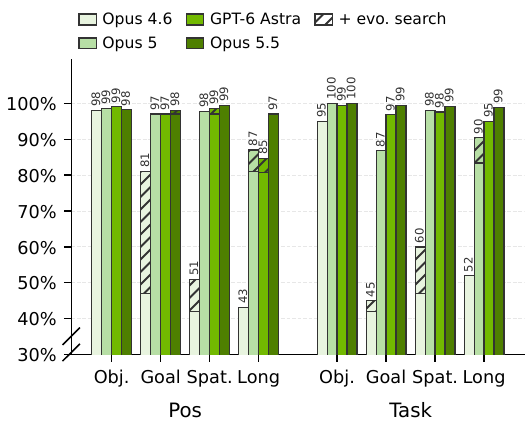}
\end{minipage}\hfill
\begin{minipage}[t]{0.44\textwidth}
\vspace{0pt}
\centering
\scriptsize
\setlength{\tabcolsep}{3pt}
\captionof{table}{\textbf{BEHAVIOR-1K.} Navigation and task success on held-out seeds.}
\label{tab:ablation-backbone-b1k}
\begin{tabular}{llcccc}
\toprule
\textbf{Task} & & \multicolumn{3}{c}{\textbf{Claude Opus}} & \textbf{GPT-6} \\
\cmidrule(lr){3-5}\cmidrule(lr){6-6}
 & & 4.6 & 5 & 5.5 & Astra \\
\midrule
\textsc{soda-can} & Nav.  & 92\%  & 100\% & 100\% & 96\% \\
                  & Task  & 88\%  & 100\% & 100\% & 96\% \\
\addlinespace
\textsc{radio}    & Nav.  & 100\% & 100\% & 100\% & 96\% \\
                  & Task  & 88\%  & 92\%  & 100\% & 96\% \\
\bottomrule
\end{tabular}

\vspace{8pt}
\captionof{table}{\textbf{Robosuite.} Success over 100 held-out trials on the three hardest tasks (all seven: Appendix~\ref{sec:appendix:model-ablation-robosuite}).}
\label{tab:ablation-backbone-robosuite}
\begin{tabular}{lcccc}
\toprule
\textbf{Task} & \multicolumn{3}{c}{\textbf{Claude Opus}} & \textbf{GPT-6} \\
\cmidrule(lr){2-4}\cmidrule(lr){5-5}
 & 4.6 & 5 & 5.5 & Astra \\
\midrule
two\_arm\_handover & 92\%  & 99\%  & 100\% & 95\%  \\
two\_arm\_lift     & 71\%  & 96\%  & 100\% & 100\% \\
nut\_assembly      & 9\%   & 98\%  & 100\% & 99\%  \\
\bottomrule
\end{tabular}
\end{minipage}
\captionof{figure}{\textbf{Model ablation on LIBERO-Pro.} Held-out success per suite and perturbation (10 tasks $\times$ 50 seeds); hatched segments are the gain from evolutionary search (Opus~4.6 on Long without it). Unlike the empty-library runs in Figure~\ref{fig:model-ablation}, these runs carry the skill library across tasks within a suite, so their Goal values under position perturbation differ. Tables~\ref{tab:ablation-backbone-b1k} and~\ref{tab:ablation-backbone-robosuite} (right) give the same comparison on BEHAVIOR-1K and Robosuite.}
\label{fig:model-ablation-pro}
\end{figure}

\smallskip\noindent\textbf{Near saturation across benchmarks.}\hspace{0.5em}With Claude Opus~5, Opus~5.5, or GPT-6 Astra (Figure~\ref{fig:model-ablation-pro}; Tables~\ref{tab:ablation-backbone-b1k} and~\ref{tab:ablation-backbone-robosuite}), iterative debugging alone reaches 96--99\% on LIBERO-Pro Object, Goal, and Spatial, against 62\% for Opus~4.6 (72\% with evolutionary search). With Opus~5.5, \name\ reaches 97.9\% on Long, 100\% on Robosuite (\texttt{nut\_assembly} rises from 9\% to 100\%), and 100\% on BEHAVIOR-1K.

\smallskip\noindent\textbf{Evolutionary search fixes the hardest tasks.}\hspace{0.5em}Stronger models leave fewer failures, so evolutionary search moves suite averages less (at most 2\% on Object, Goal, and Spatial, against up to 34\% for Opus~4.6), but it remains what repairs the tasks iterative debugging cannot: putting the mug in the microwave rises from 0\% to 42\% with Opus~5, and placing the bowl in the top drawer from 82\% to 98\% with Astra (Tables~\ref{tab:libero-per-task-object}--\ref{tab:libero-per-task-long}).

\begin{figure}[!t]
\centering
\includegraphics[width=\textwidth]{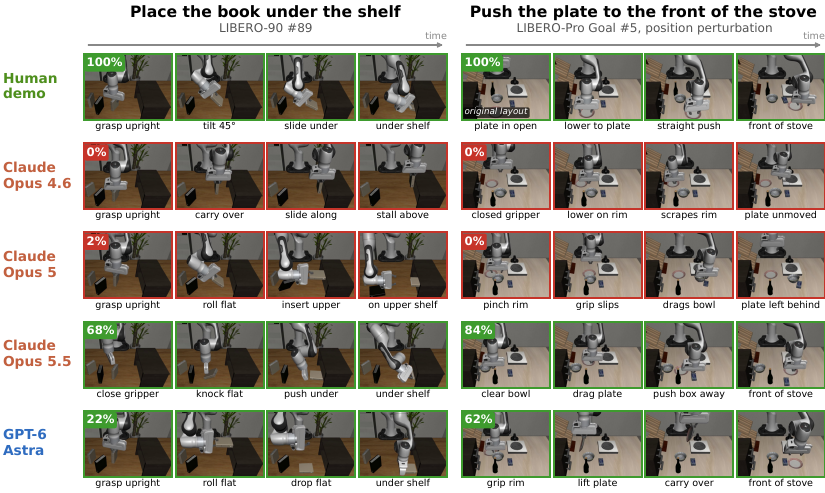}
\caption{\textbf{Case study: five solvers, two tasks.} Each row shows one episode: the human teleoperation demonstration from the original LIBERO dataset, and one held-out evaluation episode of the \name\ program written by each model. Frames run left to right in time and are labelled with the step they show. Percentages are held-out success over 50 seeds (Table~\ref{tab:case-study-sr}); the demonstrations all succeed. Green and red borders mark the shown episode's success or failure. The plate demonstration uses the original, unblocked layout. Two more tasks are in Appendix~\ref{sec:appendix:case-study}.}
\label{fig:case-study}
\end{figure}

\subsection{Case Study of the Coding Agents}
\label{sec:experiments:case-study}
Figure~\ref{fig:case-study} compares how four models and a human solve two tasks. The rows show one episode each, so they illustrate strategies, not matched trials.

\emph{Book under the shelf.} The gap under the shelf is too low for a hand holding an upright book. The human holds the book by its top edge, tilts it, and slides it in. No model finds this move. Opus~4.6 carries the book above the cabinet and stalls (0\%). Opus~5 rolls the book flat but inserts it into the compartment above the target (2\%). The two models that succeed let go of the book: Opus~5.5 knocks it flat without grasping it and pushes it under (68\%), and Astra rolls it flat, drops it, and pushes it under (22\%).

\emph{Push the plate.} In the original layout the plate lies in the open, and the human pushes it straight to the stove. The position perturbation puts a bowl in front of the plate and a box in its path. Opus~4.6 pushes without moving the plate (0\%), and Opus~5 loses its grip and drags the bowl instead (0\%). Opus~5.5 pushes the bowl and the box out of the way and then pushes the plate (84\%). Astra does not push at all: it lifts the plate over the clutter and sets it down in front of the stove (62\%).

Weaker models keep the default grasp-and-move strategy and fail at the robot's limits. Stronger models change strategy and reach the goal in different ways. Because \name\ validates each program by executing it, the same loop turns a stronger model into a more capable robot program.

\section{Related Work}
\label{sec:related}

\paragraph{Agentic robot control.}
Robot control has been studied through end-to-end vision-language-action policies and executable programs that compose perception, planning, and control APIs~\citep{ahn2022saycan, brohan2023rt2, octo2024, kim2024openvlaopensourcevisionlanguageactionmodel, black2024pi0visionlanguageactionflowmodel, mu2024robocodex, intelligence2025pi05, bjorck2025gr00t, atomvla2025, li2026roboclaw, capx2026, chen2026gap}.
Software-engineering agents provide a related write-execute-debug loop for code~\citep{jimenez2024swebench, wang2025openhands, yang2024sweagent, wang2024codeact, chen2024selfdebug, claudecode2025, openai2025codex, opencode2025}.
\name builds on the executable-program paradigm, but focuses on persistent embodied improvement: the agent freely selects primitive-level multimodal traces, writes repairs, validates, and preserves successful experiences across tasks.

\paragraph{Self-improving agents and skill libraries.}
LLM agents have been improved through open-ended memory, skill libraries, and self-evolving skill repositories~\citep{voyager2023, tziafas2024lrll, peng2024hyvin, vistawise2025, skillflow2026, uniskill2026}. Other systems use LLMs to generate rewards, curricula, environments, reset, or search candidates~\citep{eureka2023, yu2023l2r, xie2024text2reward, wang2024eurekaverse, ma2024dreureka, du2023ellm, wang2024robogen, romeraparedes2024funsearch, novikov2025alphaevolve, guo2025evoengineer, guo2026codeevolution, cao2026ksearch, xiao2026enpire}.
Unlike success-only memories, textual reflections, or reward functions for downstream policy training, \name stores validated repair knowledge extracted from attributed embodied failures. Its skills are open-ended in both content and admission: they emerge from the agent's own debugging experience, span heterogeneous categories (\S\ref{sec:method:library}), and are reused as in-context guidance for future program repair. Evolutionary search (\S\ref{sec:method:evo}) further broadens the space of executable program repairs before reusable patterns are admitted to the library.

\section{Limitations}
\label{sec:limitations}
\name\ has several important limitations.

\smallskip\noindent\textbf{Real-world autonomy.}\hspace{0.5em}Selected simulation-discovered skills support autonomous real-robot debugging on a different embodiment, and \name\ learns new skills online on a real robot, but it is not yet a fully autonomous real-world lifelong learner. In the Rubik's Cube experiment (\S\ref{sec:cube-agentic-skill-learning}), for example, a human operator still provides the initial task guidance, scrambles the cube, and resets the scene between episodes. Unlike simulation, where success checking and scene resets are cheap and programmatic, real-world deployment still requires robust success detection, safe reset, safety monitoring, and calibration maintenance. Future work should close this evaluation-and-reset loop to scale sim-to-real skill transfer across broader real-world task suites.

\smallskip\noindent\textbf{Dependence on the LLM.}\hspace{0.5em}\name\ relies on a frozen LLM to interpret multimodal traces, write program repairs, and propose evolutionary-search candidates, so its performance depends on that model. Weaker models such as GPT-6 Luna, Claude Opus~4.6, Qwen~3.8 Max, and GPT-6 Sol reach only 15--59\% on LIBERO-Pro Goal under the position perturbation, and the best open-weight model, Kimi~K3, reaches 78\% (Figure~\ref{fig:model-ablation}); our best results use proprietary frontier models.

\smallskip\noindent\textbf{Predefined robot API.}\hspace{0.5em}\name\ writes programs using a predefined API of perception, planning, and control primitives. This API makes debugging tractable and safe, but also bounds the behaviors the agent can express: if a task requires sensing, control, or interaction capabilities outside the exposed primitives, the agent must either approximate them inefficiently or rely on humans to extend the API. Future systems should study how agents can safely propose, validate, and incorporate new robot primitives.

\smallskip\noindent\textbf{Skill library management.}\hspace{0.5em}A verifier admits a skill only after it independently reproduces the skill's success on the source task (\S\ref{sec:method:library}), but this check happens once, at admission. Entries are not re-validated as the library grows, so some may later become stale, redundant, or misleading for a new task, which can hurt individual tasks in zero-shot transfer. Re-validation, pruning, and task-aware ranking at retrieval time are needed to keep the library useful at scale.

\smallskip\noindent\textbf{Compute cost.}\hspace{0.5em}The debugging and evolutionary-search loop is compute-intensive, consuming many LLM calls and simulator or robot rollouts per task. Scaling to very large task suites will require cheaper LLM inference, more sample-efficient search, or stronger mechanisms for reusing prior repairs.

Because \name\ improves as the underlying model improves (Figure~\ref{fig:model-ablation}), we expect progress in frontier and open-weight models to ease several of these limitations.

\section{Conclusion}
\label{sec:conclusion}
We present \name, a continual learning robotic system that autonomously writes and refines robot control programs, guided by multimodal execution traces and evolutionary search over candidate programs, while compounding validated repairs into a reusable skill library.
Across diverse benchmarks, \name\ outperforms existing VLA and coding-agent baselines, transfers zero-shot to unseen long-horizon tasks, and improves with the underlying model, reaching 99\% average success across all 80 LIBERO-Pro tasks and 100\% success on representative Robosuite and BEHAVIOR-1K tasks with Claude Opus~5.5.
On three real-robot tasks, skills discovered in simulation reduce real-robot programming token cost despite a different embodiment and robot API, and skills learned online let GPT-6 Astra solve a Rubik's Cube on a real bimanual robot.

\section*{Acknowledgments}

We thank Nadun Ranawaka, Jimmy Wu, Matin Furutan, Abhi Maddukuri, Yulu Gan, Matin Nikoui, and Yuqi Xie for their help with open-source support, real robot infrastructure, advice and guidance on the paper, website release, and experimental equipment. We are also grateful to the members of NVIDIA GEAR, UMich SymbioticLab, UsesysLab, UC Berkeley AUTOLab, CMU LeCAR Lab for their kind support.

\bibliographystyle{plainnat}
\bibliography{ref}

\clearpage
\appendix
\raggedbottom

\section{Skill Library Details}
\label{sec:appendix:skill-library}

This appendix expands the representative entries shown in Figure~\ref{fig:skill-library} into the skill-library taxonomy. Each entry lists: (i)~the \emph{problem} extracted from the triggering failure trace; (ii)~the \emph{when-to-apply} guard encoding the situational retrieval condition; (iii)~the validated \emph{repair snippet}; and (iv)~the origin task(s) that produced the entry.

\begin{figure}[H]
    \centering
    \includegraphics[width=0.95\textwidth]{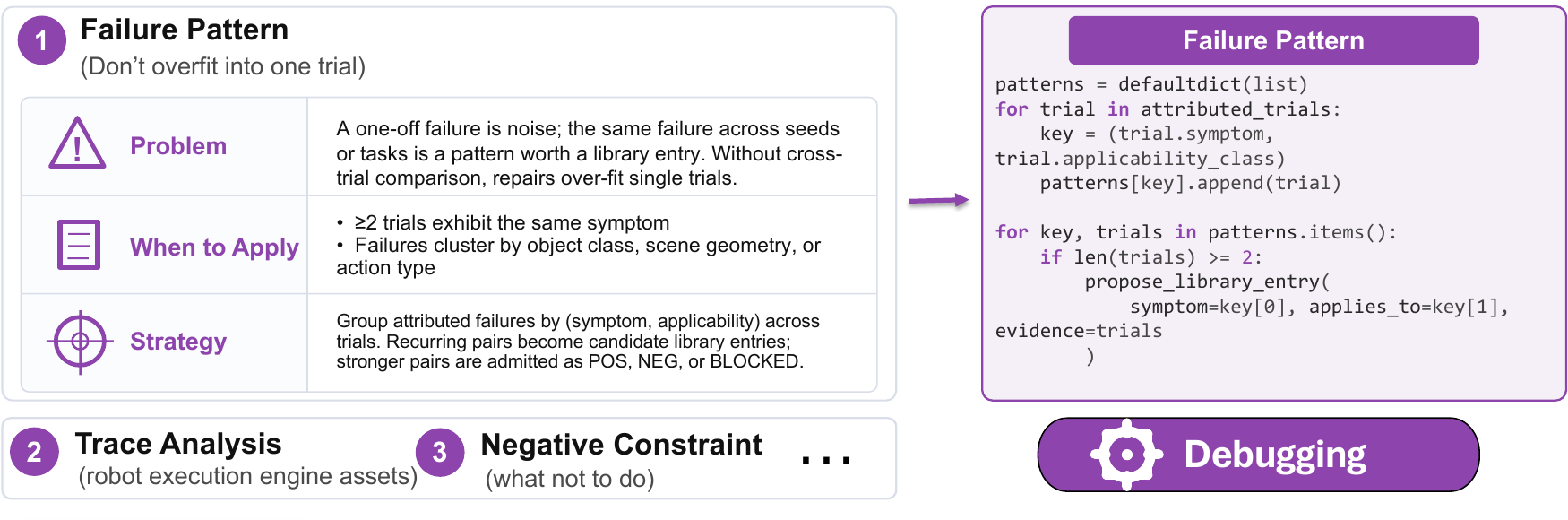}
    \caption{\textbf{Debugging skills.} Representative skill-library entries that encode reusable debugging strategies, including failure signatures, when-to-apply guards, and validated repair sketches.}
    \label{fig:appendix-skill-debugging}
\end{figure}

\begin{figure}[H]
    \centering
    \includegraphics[width=0.95\textwidth]{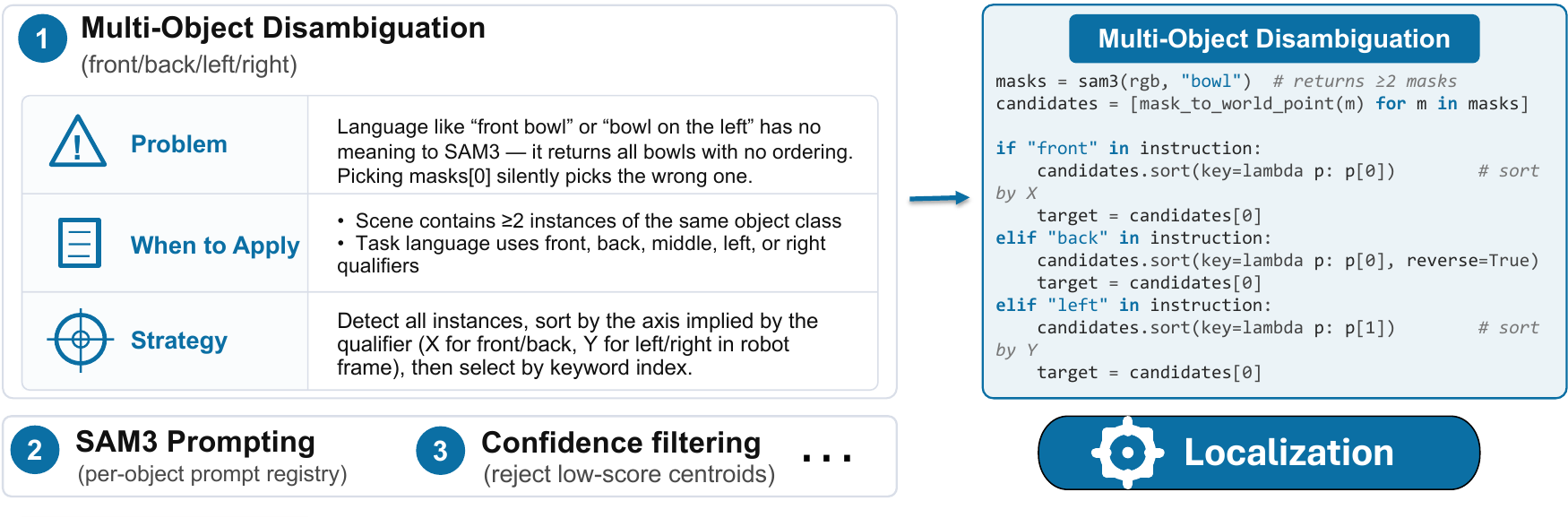}
    \caption{\textbf{Localization skills.} Representative entries for grounding ambiguous language and object references into robust perception and localization routines.}
    \label{fig:appendix-skill-localization}
\end{figure}

\begin{figure}[H]
    \centering
    \includegraphics[width=0.95\textwidth]{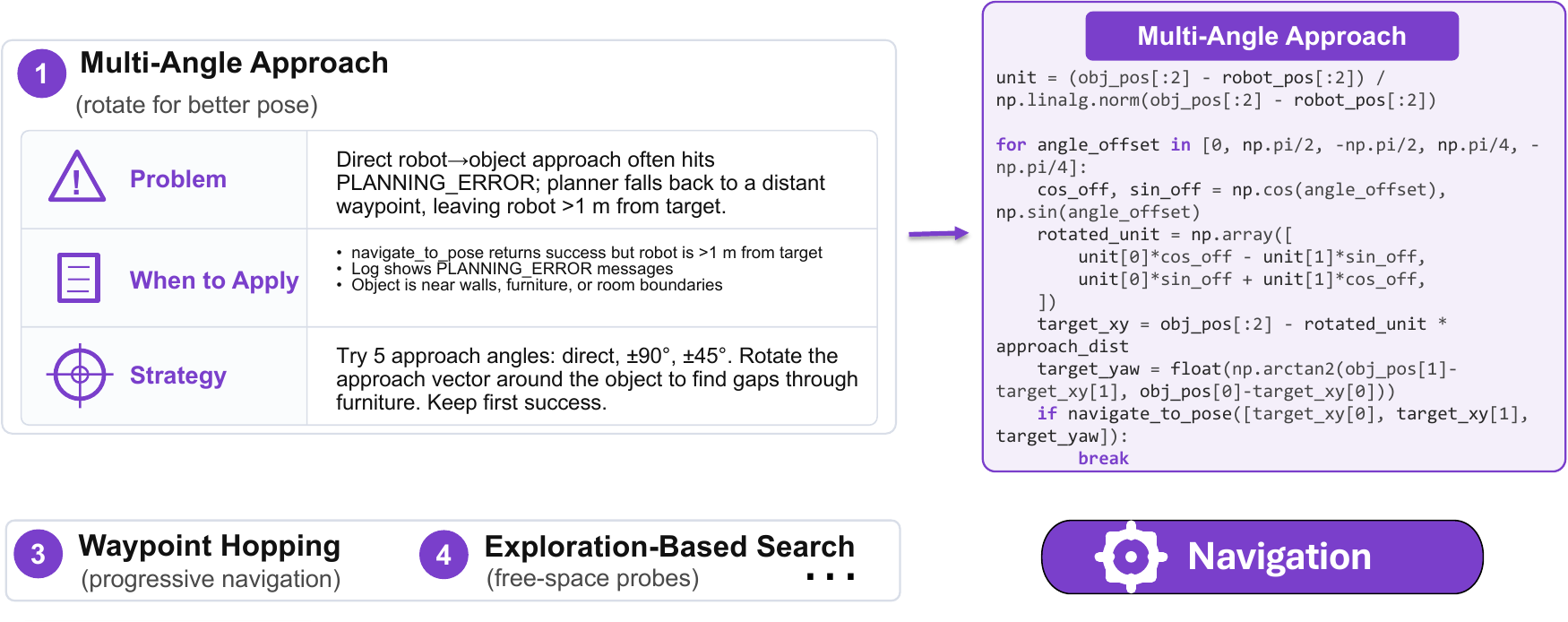}
    \caption{\textbf{Navigation skills.} Representative entries for recovering from motion-planning failures and selecting collision-aware approach poses.}
    \label{fig:appendix-skill-navigation}
\end{figure}

\begin{figure}[H]
    \centering
    \includegraphics[width=0.95\textwidth]{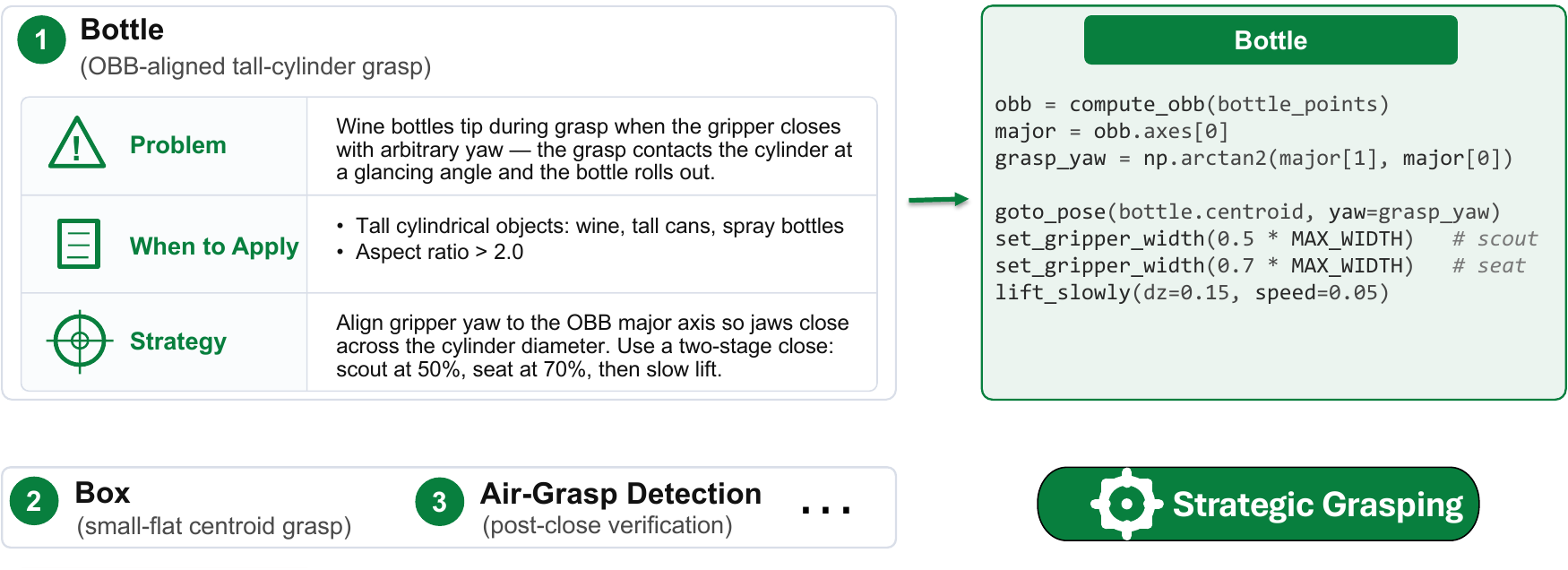}
    \caption{\textbf{Strategic grasping skills.} Representative entries for choosing task-appropriate grasp points and adapting grasp strategy to object geometry and scene context.}
    \label{fig:appendix-skill-strategic-grasping}
\end{figure}

\begin{figure}[H]
    \centering
    \includegraphics[width=0.95\textwidth]{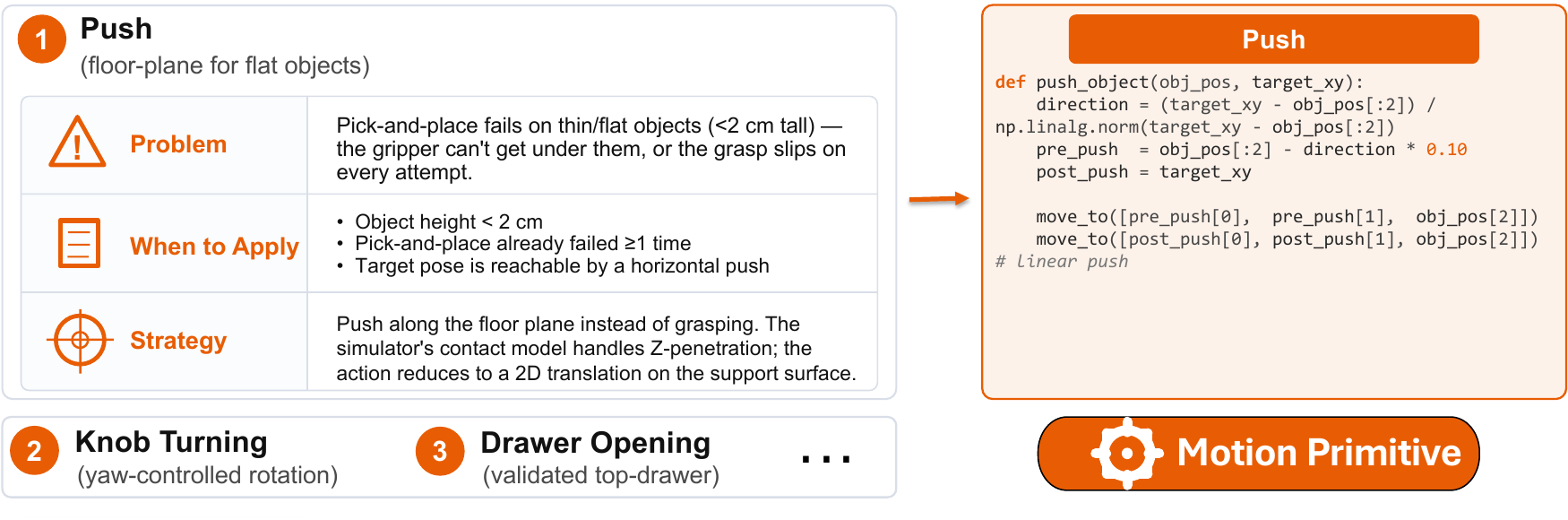}
    \caption{\textbf{Motion-primitive skills.} Representative entries for reusable low-level motion patterns, contact-rich alignment, and execution-time recovery.}
    \label{fig:appendix-skill-motion-primitive}
\end{figure}

\begin{figure}[H]
    \centering
    \includegraphics[width=0.95\textwidth]{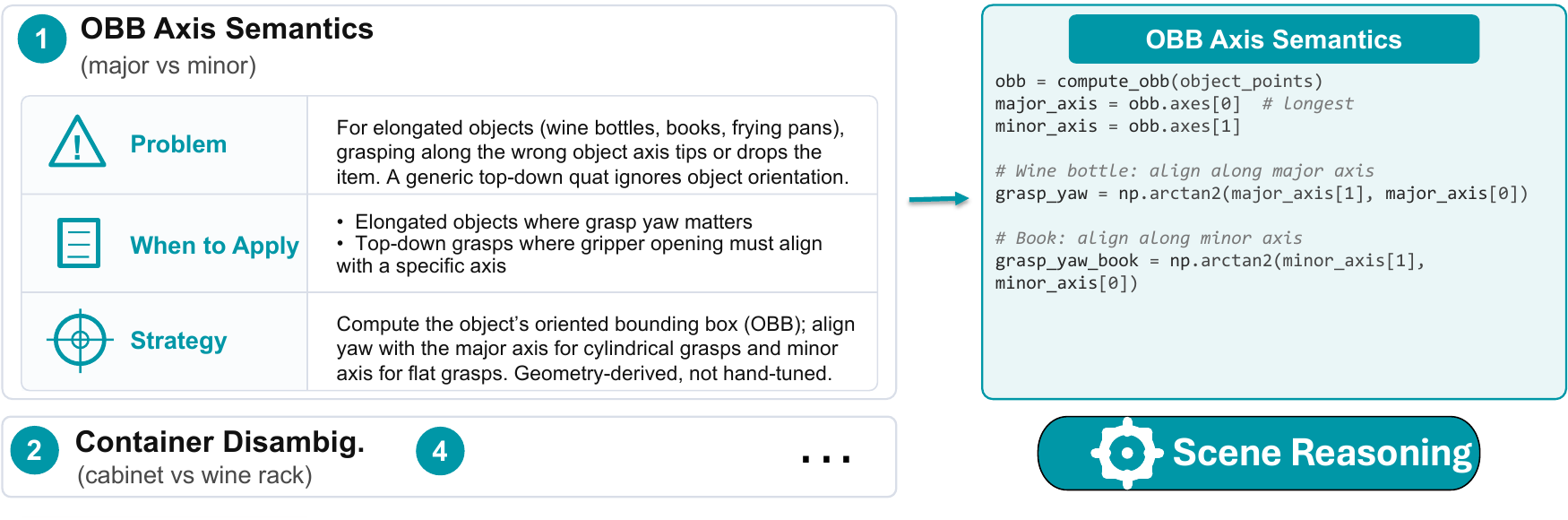}
    \caption{\textbf{Scene-reasoning skills.} Representative entries for reasoning over spatial relations, support surfaces, occlusions, and scene-level task constraints.}
    \label{fig:appendix-skill-scene-reasoning}
\end{figure}

\clearpage

\section{Main Benchmark Result Tables}
\label{sec:appendix:main-results-tables}

\subsection{Macro-Averaged Main Results}

This section reports the exact numerical values visualized in Figure~\ref{fig:main-results} of the main paper.

\begin{table}[H]
\centering
\footnotesize
\caption{\textbf{Main results on LIBERO-Pro.} Success of OpenVLA~\citep{kim2024openvlaopensourcevisionlanguageactionmodel}, $\pi_0$~\citep{black2024pi0visionlanguageactionflowmodel}, $\pi_{0.5}$~\citep{intelligence2025pi05}, CaP-Agent0~\citep{capx2026}, and \name{} (Opus 4.6) on LIBERO-Pro~\citep{zhou2025liberopro} Object, Goal, and Spatial under the position (Pos) and task (Task) perturbations, macro-averaged across 10 tasks per suite. \textbf{Overall} columns are macro-averaged across the three suites. All \name{} numbers are on the 50-seed held-out evaluation (seeds~1--50). See the per-task breakdown in Appendices~\ref{sec:appendix:ablations:swap} and~\ref{sec:appendix:ablations:task}.}
\label{tab:liberoproAVG}
\setlength{\tabcolsep}{4pt}
\begin{tabular}{lccccccccc}
\toprule
\textbf{Method} & \multicolumn{2}{c}{\textbf{Object}} & \multicolumn{2}{c}{\textbf{Goal}} & \multicolumn{2}{c}{\textbf{Spatial}} & \multicolumn{3}{c}{\textbf{Overall}} \\
\cmidrule(lr){2-3} \cmidrule(lr){4-5} \cmidrule(lr){6-7} \cmidrule(lr){8-10}
 & Pos & Task & Pos & Task & Pos & Task & Pos & Task & All \\
\midrule
OpenVLA         & 0.0\% & 0.0\% & 0.0\% & 0.0\% & 0.0\% & 0.0\% & 0.0\% & 0.0\% & 0.0\% \\
$\pi_0$         & 0.0\% & 0.0\% & 0.0\% & 0.0\% & 0.0\% & 0.0\% & 0.0\% & 0.0\% & 0.0\% \\
$\pi_{0.5}$     & 17.0\% & 1.0\% & 38.0\% & 0.0\% & 20.0\% & 1.0\% & 25.0\% & 1.0\% & 13.0\% \\
CaP-Agent0     & 22.0\% & 18.0\% & 26.0\% & 17.0\% & 12.0\% & 14.0\% & 20.0\% & 16.0\% & 18.0\% \\
\textbf{\name}  & \textbf{98.0\%} & \textbf{95.0\%} & \textbf{81.0\%} & \textbf{45.0\%} & \textbf{51.0\%} & \textbf{60.0\%} & \textbf{77.0\%} & \textbf{67.0\%} & \textbf{72.0\%} \\
\bottomrule
\end{tabular}
\end{table}

\begin{table}[H]
\centering
\footnotesize
\caption{\textbf{Main results on Robosuite.} Success of CaP-Agent0~\citep{capx2026} and \name{} (Opus 4.6) on 7 Robosuite~\citep{zhu2020robosuite} manipulation tasks (100-trial held-out evaluation). \textbf{Mean} is the unweighted mean across all 7 tasks.}
\label{tab:robosuite}
\setlength{\tabcolsep}{12pt} %
\begin{tabular}{lcc}
\toprule
\textbf{Task} & \textbf{CaP-Agent0} & \textbf{\name} \\
\midrule
cube\_lift        & \textbf{97\%} & \textbf{97\%} \\
cube\_stack       & 98\%          & \textbf{99\%} \\
cube\_restack     & 89\%          & \textbf{100\%} \\
spill\_wipe       & \textbf{100\%} & 99\%          \\
two\_arm\_handover& 20\%          & \textbf{92\%} \\
two\_arm\_lift    & \textbf{74\%} & 71\%          \\
nut\_assembly     & 0\%          & \textbf{9\%} \\
\midrule
\textit{Mean}  & \textit{68.0\%}    & \textit{81.0\%} \\
\bottomrule
\end{tabular}
\end{table}

\begin{table}[H]
\centering
\footnotesize
\caption{\textbf{Main results on BEHAVIOR-1K.} Navigation (Nav) and task (Task) success on two representative BEHAVIOR-1K~\citep{behavior1k2023} household tasks, Soda Can pick-up and Radio pick-up, reported separately. All numbers are on the 25-seed held-out evaluation (seeds 1--25) using Opus 4.6. \name{} runs interactive block-by-block generation with the skill library accumulated on seeds 26--35.}
\label{tab:b1k}
\setlength{\tabcolsep}{5pt}
\begin{tabular}{lcccccc}
\toprule
\textbf{Task} & \multicolumn{2}{c}{\textbf{Human}} & \multicolumn{2}{c}{\textbf{CaP-Agent0}} & \multicolumn{2}{c}{\textbf{\name}} \\
\cmidrule(lr){2-3} \cmidrule(lr){4-5} \cmidrule(lr){6-7}
 & Nav & Task & Nav & Task & Nav & Task \\
\midrule
Soda Can pick-up  & 80\% & 72\% & 84\% & 72\% & \textbf{92\%} & \textbf{88\%} \\
Radio pick-up     & 88\% & 36\% & 80\% & 56\% & \textbf{100\%} & \textbf{88\%} \\
\bottomrule
\end{tabular}
\end{table}

\clearpage
\section{LIBERO-Pro Long Zero-Shot Transfer}
\label{sec:appendix:libero-long-pro-zero-shot}

\subsection{Macro-Averaged Main Results}
\label{sec:appendix:libero-long-pro-zero-shot:macro-average}
\begin{table}[H]
\centering
\scriptsize
\caption{\textbf{Zero-shot transfer to LIBERO-Pro Long.} Macro-averaged success of the \name{} skill library, accumulated on LIBERO-90, on LIBERO-Pro Long~\citep{zhou2025liberopro} using Opus 4.6.$N$ denotes the number of LIBERO-90 tasks whose repair skills seed the library; no additional debugging is performed at test time. Pos and Task denote the position and task perturbations, 10 tasks each.}
\label{tab:transfer-summary}
\setlength{\tabcolsep}{4pt}
\begin{tabular}{lccc}
\toprule
\textbf{Method} & \multicolumn{3}{c}{\textbf{LIBERO-Pro Long Zero-Shot Transfer}} \\
\cmidrule(lr){2-4}
 & \textbf{Pos} & \textbf{Task} & \textbf{Overall} \\
\midrule
OpenVLA                                   & 0.0\%          & 0.0\%          & 0.0\% \\
$\pi_0$                                   & 0.0\%          & 0.0\%          & 0.0\% \\
$\pi_{0.5}$                               & 8.0\%          & 1.0\%          & 5.0\% \\
CaP-Agent0                               & 5.2\%         & 2.4\%         & 3.8\% \\
\midrule
\name~($N{=}0$, zero-shot)                & 0.0\%          & 9.4\%         & 4.7\% \\
\name~($N{=}25$, zero-shot)               & 5.6\%         & 21.8\%         & 13.7\% \\
\name~($N{=}50$, zero-shot)               & 13.8\%         & 29.2\%         & 21.5\% \\
\name~($N{=}90$, zero-shot)               & \textbf{22.6\%} & \textbf{38.3\%} & \textbf{30.5\%} \\
\bottomrule
\end{tabular}
\end{table}

\subsection{Per-Task Breakdown}
\label{sec:appendix:libero-long-pro-zero-shot:per-task}
\begin{table}[H]
\centering
\scriptsize
\caption{\textbf{Per-task zero-shot transfer to LIBERO-Pro Long.} As in Table~\ref{tab:transfer-summary}, per task and snapshot size $N$ (seeds 1--50, $n{=}50$).
  \textbf{Pos}: position perturbation.
  \textbf{Task}: task perturbation.}
\label{tab:libero-long-scaling}
\setlength{\tabcolsep}{3pt}
\begin{tabular}{>{\scriptsize}lcccc cccc}
\toprule
& \multicolumn{4}{c}{\textbf{Pos}} & \multicolumn{4}{c}{\textbf{Task}} \\
\cmidrule(lr){2-5}\cmidrule(lr){6-9}
\textbf{Task} & $N{=}0$ & $N{=}25$ & $N{=}50$ & $N{=}90$
              & $N{=}0$ & $N{=}25$ & $N{=}50$ & $N{=}90$ \\
\midrule
Stove + moka pot      & 0\% & 30\% & 84\% & \textbf{100\%} & 0\% & 2\% & 68\% & 26\% \\
Bowl in drawer        & 0\% & 0\% & 0\% & 0\%          & 0\% & 0\% & 0\% & 0\% \\
Mug in microwave      & 0\% & 0\% & 0\% & 0\%          & 0\% & 0\% & 0\% & 0\% \\
Both mokas on stove   & 0\% & 0\% & 0\% & 0\%          & 0\% & 0\% & 84\% & \textbf{100\%} \\
Soup + cream cheese   & 0\% & 12\% & 2\% & 0\%          & 0\% & 68\% & 60\% & 70\% \\
Soup + tomato sauce   & 0\% & 14\% & 20\% & 0\%          & 78\% & 68\% & 2\% & 70\% \\
Cream cheese + butter & 0\% & 0\% & 6\% & \textbf{94\%} & 0\% & 0\% & 34\% & 86\% \\
Mug on two plates     & 0\% & 0\% & 16\% & 32\%          & 16\% & 28\% & 8\% & 0\% \\
Mug + pudding         & 0\% & 0\% & 6\% & 0\%          & 0\% & 0\% & 8\% & 4\% \\
Book in caddy         & 0\% & 0\% & 4\% & 0\%          & 0\% & 52\% & 28\% & 26\% \\
\midrule
\textit{Mean}         & \textit{0.0\%} & \textit{6.0\%} & \textit{14.0\%} & \textit{23.0\%}
                      & \textit{9.0\%} & \textit{22.0\%} & \textit{29.0\%} & \textit{38.0\%} \\
\bottomrule
\end{tabular}
\end{table}

\clearpage

\section{Harness for an Agentic System 2 on Real Robots}
\label{app:aspire-system2-architecture}

\begin{figure}[hbp]
\centering
\includegraphics[width=0.8\textwidth]{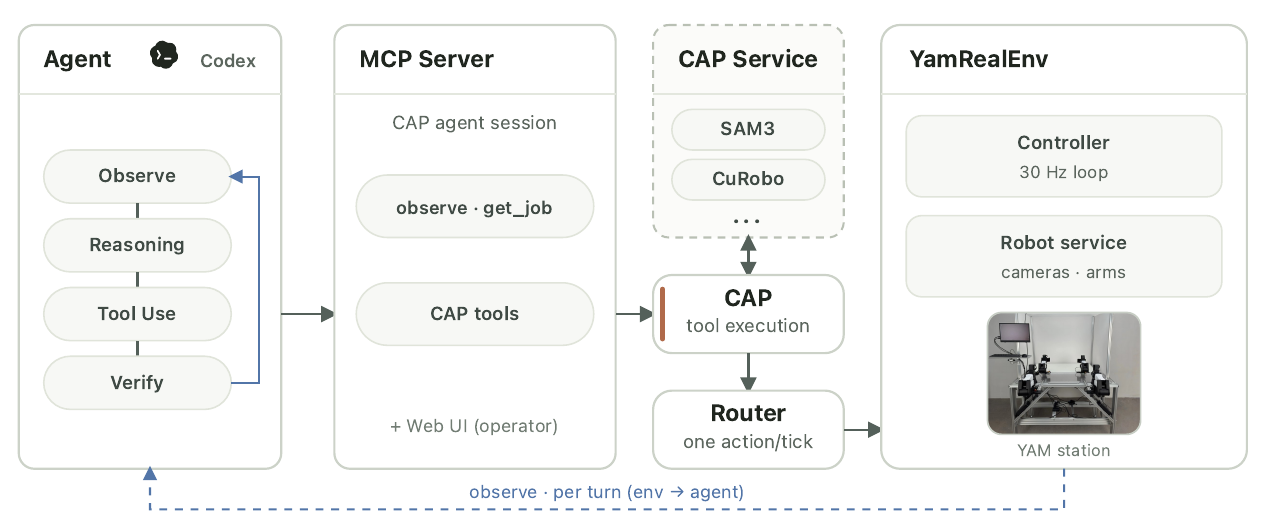}
\caption{\textbf{System design for the agentic workflow.} The coding agent runs an observe--reason--act--verify loop. It may call observation tools to acquire four-view images or target object pose estimates. CaP tools provide action primitives such as end-effector control.}
\label{fig:sys2-design}
\end{figure}

Apart from the execution of the open-loop program, we consider extending the harness to support the closed-loop interaction of coding agents, as shown in Figure~\ref{fig:sys2-design}. The agent serves as a System 2 connected to a persistent Code-as-Policy (CAP) runtime through the Model Context Protocol (MCP). The agent interprets tasks, plans, selects actions, and evaluates outcomes. The runtime comprises an \texttt{MCP bridge}, a shared \texttt{session manager}, \texttt{CAP API} with perception and planning providers, and the \text{real YAM environment} that owns arm services. The \texttt{MCP bridge} serves a Streamable HTTP endpoint with an explicit allowlist of tools. Requests are validated against JSON schemas before dispatch. Tool availability reflects the configured providers; discovery also reports provider identifiers and source hashes for packaged CAP programs. The interface exposes the following classes of capabilities:

\begin{itemize}
    \item \textbf{Discovery and monitoring.}
    \texttt{get\_capability\_graph} describes capabilities;
    \texttt{get\_job} reports the current or most recent job; and
    \texttt{read\_operator\_messages} retrieves queued instructions.

    \item \textbf{Observation and perception.}
    \texttt{observe} returns state, camera images, and freshness metadata.
    \texttt{cap\_get\_robot\_state} returns joint, gripper, and end-effector state.
    \texttt{cap\_localize\_object} estimates geometry from a text query and
    calibrated RGB-D.

    \item \textbf{Motion and grasping.}
    \texttt{cap\_move\_to\_pose} commands an absolute end-effector pose that the arm will be tracking; 
    \texttt{cap\_move\_to\_joints} directly commands six joint angles in radians;
    \texttt{cap\_move\_synchronized} coordinates targets for both arms; and
    \texttt{cap\_set\_gripper} commands one gripper with a normalized opening
    in $[0,1]$, from closed to open.
\end{itemize}

For the basic action tool \texttt{cap\_move\_to\_pose}, the agent must choose the Cartesian target position. Internally, the tool uses Mink for inverse kinematics and cuRobo for trajectory planning. Physical operations run as serialized background jobs, with concurrent actuation requests rejected. MCP action calls wait for their job to finish and return structured status, measured terminal state, errors and diagnostics, and camera images as native MCP image content. Observations include timestamps, frame identifiers, acquisition age, and stale or missing flags. Control-level success only confirms the commanded motion; the agent judges task completion from physical evidence.

On the deployment end, \texttt{RealYamEnv} owns camera acquisition and connects to the arm services through RPC. The CAP adapter routes actuation through one control worker.
Trajectories are interpolated in time and streamed as joint and gripper position
targets to the arm services and CAN motor drivers. Station profiles supply
hardware endpoints, camera identities and calibration, robot models, joint
limits, and home targets. Actuation requires explicit enablement at server startup.

\paragraph{Agent Sessions and Closed-Loop Workflow}
A fresh agent session is launched through the Codex CLI with the agent operating
guide and operator task. Repository instructions define execution rules;
tool-use and task-specific skills supply action semantics and manipulation
strategies. The conversation retains the goal, plan, observations, and unresolved
subgoals. The robot runtime persists independently, allowing a fresh agent to
attach without recreating the environment or reopening cameras. The session instructions specify the agent loop: 
1) \textbf{Initialize.} Read the applicable instructions and discover the available tools. A motion-enabled fresh session invokes Home once, checks its terminal result, measured joint residuals, gripper state, and camera freshness, and reports readiness. Observation-only startup performs readiness checks without motion; 
2) \textbf{Observe and reason.} Inspect the scene, obtain calibrated object geometry when needed, and identify the next subgoal and its completion evidence. 
3) \textbf{Act.} Select a registered motion or gripper tool, ground its targets in the available observations, and wait for its terminal result before issuing a dependent action.
4) \textbf{Verify.} Evaluate returned state and images against the subgoal. Reuse sufficient fresh evidence; request another observation or localization when evidence is stale, ambiguous, or invalidated by motion. 

\begin{figure}[hbp]
\centering
\includegraphics[width=0.8\textwidth]{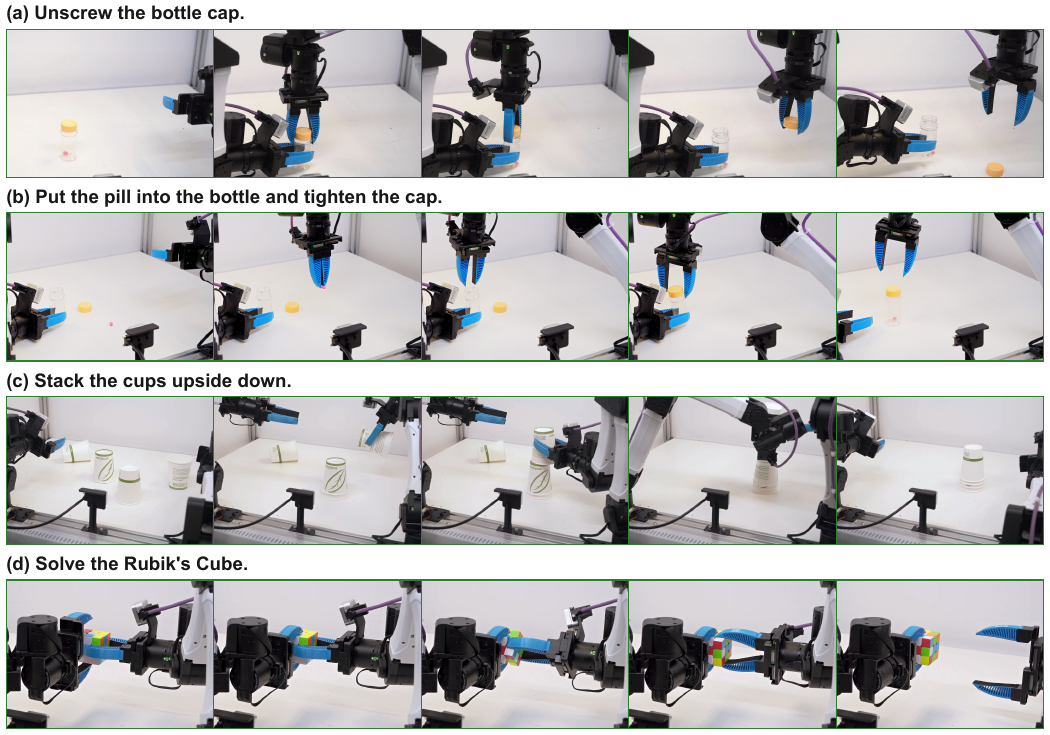}
\caption{\textbf{Real robot experiments on system-2.} We provide example for real robot tasks on a bi-manual YAM platform that require atomic manipulation skills including pick-and-place, handover, re-orientation and fine-grained adjustments.}
\label{fig:real-tasks-traj}
\end{figure}

With this System 2 harness, we show that \textbf{\texttt{GPT-6 Astra}} can perform contact-rich, long-horizon manipulation tasks as demonstrated in \ref{fig:real-tasks-traj}. Specifically, we task solving \textbf{Rubik's Cube} from randomly initialized cube states. Human operator would provide an heuristic context documents (general rules and potential failure cases, in text form), and the agent explores reliable skills or motion patterns such as cube re-orientation and face turning. The agent first rotates the cube to expose all faces and perceive its state, then calls the Kociemba solver to obtain a solution, and then executes the plan while updating its memory of the cube state.

\clearpage

\section{Ablation Details}
\label{sec:appendix:ablations}

\subsection{LIBERO-Pro Position Perturbation Per-Task Breakdown}
\label{sec:appendix:ablations:swap}

\begin{table}[H]
\centering
\scriptsize
\caption{\textbf{Per-task ablation on LIBERO-Pro, position perturbation, Opus 4.6.} Success on seeds 1--50 ($n{=}50$).
  \textbf{w/o Engine \& Evo}: \name{} without the robot execution engine and evolutionary search, i.e., zero-shot Claude Opus~4.6 with 15 example programs.
  \textbf{w/o Evo}: \name{} without evolutionary search, i.e., the program repaired with the robot execution engine and skill library.
  \textbf{Evo.\ search}: raw evolutionary search result (best validated candidate), or ``--'' if evolutionary search was not run for that task.
  \textbf{\name}: per-task winner between the repaired program and evolutionary search, selected on a held-out validation set (seeds 66--80).}
\label{tab:libero-pro-ablation-pos}
\setlength{\tabcolsep}{2.5pt}
\begin{tabular}{>{\scriptsize}lcccc}
\toprule
\textbf{Task} & \textbf{~w/o Engine \& Evo} & \textbf{~w/o Evo} & \textbf{Evo.\ search} & \textbf{\name} \\
\midrule
\multicolumn{5}{l}{\textit{LIBERO-Pro Object}} \\[1pt]
Alphabet soup               & 0\% & \textbf{100\%} & -- & \textbf{100\%} \\
BBQ sauce                   & 6\% & \textbf{100\%} & -- & \textbf{100\%} \\
Butter                      & 0\% & \textbf{100\%} & -- & \textbf{100\%} \\
Chocolate pudding           & 0\% & \textbf{100\%} & -- & \textbf{100\%} \\
Cream cheese                & 92\% & \textbf{100\%} & -- & \textbf{100\%} \\
Ketchup                     & 2\% & \textbf{92\%} & -- & \textbf{92\%} \\
Milk                        & 6\% & \textbf{92\%} & -- & \textbf{92\%} \\
Orange juice                & 46\% & \textbf{100\%} & -- & \textbf{100\%} \\
Salad dressing              & 18\% & \textbf{96\%} & -- & \textbf{96\%} \\
Tomato sauce                & 66\% & \textbf{98\%} & -- & \textbf{98\%} \\[1pt]
\textit{Mean}               & \textit{24.0\%} & \textit{98.0\%} & \textit{--} & \textit{98.0\%} \\
\midrule
\multicolumn{5}{l}{\textit{LIBERO-Pro Goal}} \\[1pt]
Open middle drawer          & 28\% & 2\% & 72\% & \textbf{72\%} \\
Open top drawer+bowl        & 0\% & \textbf{86\%} & -- & \textbf{86\%} \\
Push plate to stove         & 0\% & 0\% & 0\% & 0\% \\
Bowl on plate               & 4\% & 36\% & 92\% & \textbf{92\%} \\
Bowl on stove               & 16\% & 56\% & 96\% & \textbf{96\%} \\
Bowl on cabinet top         & 72\% & 74\% & 100\% & \textbf{100\%} \\
Cream cheese in bowl        & 0\% & \textbf{74\%} & 92\% & \textbf{74\%} \\
Wine on rack                & 6\% & 46\% & 98\% & \textbf{98\%} \\
Wine on cabinet top         & \textbf{100\%} & 24\% & 100\% & \textbf{100\%} \\
Turn on stove               & 40\% & 68\% & 90\% & \textbf{90\%} \\[1pt]
\textit{Mean}               & \textit{27.0\%} & \textit{47.0\%} & \textit{82.0\%}$^{(n{=}9)}$ & \textit{81.0\%} \\
\midrule
\multicolumn{5}{l}{\textit{LIBERO-Pro Spatial}} \\[1pt]
Betw.\ plate+ramekin        & 2\% & 10\% & 94\% & \textbf{94\%} \\
From table center           & 0\% & \textbf{32\%} & 82\% & \textbf{32\%} \\
In top drawer               & 2\% & \textbf{22\%} & 10\% & 10\% \\
Next to cookie box          & 2\% & \textbf{16\%} & 12\% & \textbf{16\%} \\
Next to plate               & 8\% & \textbf{100\%} & 100\% & \textbf{100\%} \\
Next to ramekin             & 0\% & \textbf{20\%} & 40\% & \textbf{20\%} \\
On cookie box               & 0\% & \textbf{32\%} & 24\% & 24\% \\
On ramekin                  & 58\% & 42\% & 94\% & \textbf{94\%} \\
On stove                    & 4\% & \textbf{86\%} & -- & \textbf{86\%} \\
On wooden cabinet           & 10\% & \textbf{56\%} & 34\% & 34\% \\[1pt]
\textit{Mean}               & \textit{9.0\%} & \textit{42.0\%} & \textit{54.0\%}$^{(n{=}9)}$ & \textit{51.0\%} \\
\midrule
\textit{Overall mean}       & \textit{20.0\%} & \textit{62.0\%} & \textit{68.0\%}$^{(n{=}18)}$ & \textit{77.0\%} \\
\bottomrule
\end{tabular}
\end{table}

\subsection{LIBERO-Pro Task Perturbation Per-Task Breakdown}
\label{sec:appendix:ablations:task}

\begin{table}[H]
\centering
\scriptsize
\caption{\textbf{Per-task ablation on LIBERO-Pro, task perturbation, Opus 4.6.} As in Table~\ref{tab:libero-pro-ablation-pos}, for the task perturbation.}
\label{tab:libero-pro-ablation-task}
\setlength{\tabcolsep}{2.5pt}
\begin{tabular}{>{\scriptsize}lcccc}
\toprule
\textbf{Task} & \textbf{~w/o Engine \& Evo} & \textbf{~w/o Evo} & \textbf{Evo.\ search} & \textbf{\name} \\
\midrule
\multicolumn{5}{l}{\textit{LIBERO-Pro Object}} \\[1pt]
Alphabet soup               & 0\% & \textbf{100\%} & -- & \textbf{100\%} \\
BBQ sauce                   & 88\% & \textbf{100\%} & -- & \textbf{100\%} \\
Butter                      & 78\% & \textbf{98\%} & -- & \textbf{98\%} \\
Chocolate pudding           & 0\% & \textbf{100\%} & -- & \textbf{100\%} \\
Cream cheese                & 0\% & \textbf{72\%} & 74\% & \textbf{72\%} \\
Ketchup                     & 0\% & \textbf{96\%} & -- & \textbf{96\%} \\
Milk                        & 0\% & \textbf{100\%} & -- & \textbf{100\%} \\
Orange juice                & 0\% & \textbf{100\%} & -- & \textbf{100\%} \\
Salad dressing              & 0\% & \textbf{92\%} & -- & \textbf{92\%} \\
Tomato sauce                & 4\% & \textbf{96\%} & -- & \textbf{96\%} \\[1pt]
\textit{Mean}               & \textit{17.0\%} & \textit{95.0\%} & \textit{--} & \textit{95.0\%} \\
\midrule
\multicolumn{5}{l}{\textit{LIBERO-Pro Goal}} \\[1pt]
Open middle drawer          & 0\% & 0\% & 0\% & 0\% \\
Open top drawer+bowl        & 0\% & \textbf{50\%} & 46\% & 46\% \\
Push plate to stove         & 26\% & 58\% & 76\% & \textbf{76\%} \\
Bowl on plate               & 0\% & \textbf{88\%} & 82\% & \textbf{88\%} \\
Bowl on stove               & 0\% & 0\% & 0\% & 0\% \\
Bowl on cabinet top         & 0\% & 0\% & 4\% & \textbf{4\%} \\
Cream cheese in bowl        & 0\% & \textbf{82\%} & 74\% & 74\% \\
Wine on rack                & 0\% & 74\% & 86\% & \textbf{86\%} \\
Wine on cabinet top         & 0\% & \textbf{14\%} & 72\% & \textbf{14\%} \\
Turn on stove               & 24\% & 52\% & 66\% & \textbf{66\%} \\[1pt]
\textit{Mean}               & \textit{5.0\%} & \textit{42.0\%} & \textit{51.0\%}$^{(n{=}10)}$ & \textit{45.0\%} \\
\midrule
\multicolumn{5}{l}{\textit{LIBERO-Pro Spatial}} \\[1pt]
Betw.\ plate+ramekin        & \textbf{14\%} & 8\% & 84\% & 8\% \\
From table center           & 26\% & 46\% & 64\% & \textbf{64\%} \\
In top drawer               & 4\% & \textbf{96\%} & -- & \textbf{96\%} \\
Next to cookie box          & 0\% & \textbf{94\%} & -- & \textbf{94\%} \\
Next to plate               & 8\% & \textbf{70\%} & 92\% & \textbf{70\%} \\
Next to ramekin             & 0\% & \textbf{60\%} & 42\% & \textbf{60\%} \\
On cookie box               & 0\% & 6\% & 70\% & \textbf{70\%} \\
On ramekin                  & 20\% & 10\% & 46\% & \textbf{46\%} \\
On stove                    & 0\% & 32\% & 42\% & \textbf{42\%} \\
On wooden cabinet           & 13\% & \textbf{46\%} & 98\% & \textbf{46\%} \\[1pt]
\textit{Mean}               & \textit{4.0\%} & \textit{47.0\%} & \textit{67.0\%}$^{(n{=}8)}$ & \textit{60.0\%} \\
\midrule
\textit{Overall mean}       & \textit{9.0\%} & \textit{61.0\%} & \textit{59.0\%}$^{(n{=}19)}$ & \textit{67.0\%} \\
\bottomrule
\end{tabular}
\end{table}

\subsection{Evolutionary Search Progress Per-Task Breakdown}
\label{sec:appendix:ablations:evo-search}
\begin{table}[H]
\centering
\scriptsize
\caption{\textbf{Evolutionary search progress.} Held-out success (seeds 1--50) of the best
candidate at each iteration on selected LIBERO-Pro tasks. Opus 4.6. Blank cells indicate that the search terminated early.}
\label{tab:evo-progress}
\setlength{\tabcolsep}{4pt}
\begin{tabular}{llccccc}
\toprule
\textbf{Suite} & \textbf{Task} & \textbf{Iter 0} & \textbf{Iter 1} & \textbf{Iter 2} & \textbf{Iter 3} & \textbf{Iter 4} \\
\midrule
Goal (Pos)    & Bowl $\to$ plate         & 62\% & 60\% & 60\% & 18\% & \textbf{86\%} \\
Goal (Pos)    & Wine bottle $\to$ rack   & 40\% & \textbf{76\%} & 74\% &      &      \\
Goal (Pos)    & Bowl $\to$ stove         & 62\% & \textbf{82\%} &      &      &      \\
Goal (Task)   & Push plate $\to$ stove   & 0\% & \textbf{80\%} &      &      &      \\
Spatial (Pos) & Bowl on cabinet $\to$ plate & 16\% & 38\% & 38\% & \textbf{46\%} & 30\% \\
Spatial (Pos) & Bowl next to cookie $\to$ plate & 16\% & 20\% & 4\% & 4\% & \textbf{40\%} \\
Spatial (Task) & Bowl on cookie $\to$ plate & 26\% & 4\% & \textbf{70\%} &      &      \\
Spatial (Task) & Bowl on ramekin $\to$ plate & 2\% & 0\% & 22\% & \textbf{36\%} & 18\% \\
\bottomrule
\end{tabular}
\end{table}

\subsection{Model Ablation on LIBERO-Pro With and Without Evolutionary Search}
\label{sec:appendix:backbone-libero}

\begin{table}[H]
\centering
\footnotesize
\caption{\textbf{Per-task model ablation on LIBERO-Pro Object.} Success over 50 held-out seeds with (w/) and without (w/o) evolutionary search (evo). Tasks are labelled by a shortened form of the instruction the policy receives (the perturbed one under the task perturbation). Object tasks pick the named item and place it in the basket. Tasks not selected for evolutionary search keep their w/o evo value in the w/ evo column. Bold marks tasks where evolutionary search changed the held-out result.}
\label{tab:libero-per-task-object}
\setlength{\tabcolsep}{5pt}
\begin{tabular}{lcccccc}
\toprule
\textbf{Task} & \multicolumn{2}{c}{\textbf{Claude Opus~5}} & \multicolumn{2}{c}{\textbf{Claude Opus~5.5}} & \multicolumn{2}{c}{\textbf{GPT-6 Astra}} \\
\cmidrule(lr){2-3}\cmidrule(lr){4-5}\cmidrule(lr){6-7}
 & w/o evo & w/ evo & w/o evo & w/ evo & w/o evo & w/ evo \\
\midrule
\multicolumn{7}{l}{\emph{Position perturbation}} \\
0~~Alphabet soup & 100\% & 100\% & 100\% & 100\% & 100\% & 100\% \\
1~~Cream cheese & 100\% & 100\% & 100\% & 100\% & 100\% & 100\% \\
2~~Salad dressing & 100\% & 100\% & 100\% & 100\% & 100\% & 100\% \\
3~~BBQ sauce & 92\% & 92\% & 88\% & 88\% & 90\% & 90\% \\
4~~Ketchup & 100\% & 100\% & 100\% & 100\% & 100\% & 100\% \\
5~~Tomato sauce & 100\% & 100\% & 100\% & 100\% & 100\% & 100\% \\
6~~Butter & 100\% & 100\% & 100\% & 100\% & 100\% & 100\% \\
7~~Milk & 94\% & 94\% & 94\% & 94\% & 100\% & 100\% \\
8~~Chocolate pudding & 100\% & 100\% & 100\% & 100\% & 100\% & 100\% \\
9~~Orange juice & 100\% & 100\% & 100\% & 100\% & 100\% & 100\% \\
\cmidrule(lr){2-7}
\emph{Average} & 98.6\% & 98.6\% & 98.2\% & 98.2\% & 99.0\% & 99.0\% \\
\midrule
\multicolumn{7}{l}{\emph{Task perturbation}} \\
0~~Cream cheese & 100\% & 100\% & 100\% & 100\% & 100\% & 100\% \\
1~~Alphabet soup & 100\% & 100\% & 100\% & 100\% & 100\% & 100\% \\
2~~Tomato sauce & 100\% & 100\% & 100\% & 100\% & 100\% & 100\% \\
3~~Ketchup & 100\% & 100\% & 100\% & 100\% & 100\% & 100\% \\
4~~Milk & 100\% & 100\% & 100\% & 100\% & 94\% & 94\% \\
5~~BBQ sauce & 100\% & 100\% & 100\% & 100\% & 100\% & 100\% \\
6~~Orange juice & 100\% & 100\% & 100\% & 100\% & 100\% & 100\% \\
7~~Butter & 100\% & 100\% & 100\% & 100\% & 100\% & 100\% \\
8~~Salad dressing & 100\% & 100\% & 100\% & 100\% & 100\% & 100\% \\
9~~Chocolate pudding & 100\% & 100\% & 100\% & 100\% & 100\% & 100\% \\
\cmidrule(lr){2-7}
\emph{Average} & 100.0\% & 100.0\% & 100.0\% & 100.0\% & 99.4\% & 99.4\% \\
\bottomrule
\end{tabular}
\end{table}

\begin{table}[H]
\centering
\footnotesize
\caption{\textbf{Per-task model ablation on LIBERO-Pro Goal.} As in Table~\ref{tab:libero-per-task-object}, for LIBERO-Pro Goal.}
\label{tab:libero-per-task-goal}
\setlength{\tabcolsep}{5pt}
\begin{tabular}{lcccccc}

\toprule
\textbf{Task} & \multicolumn{2}{c}{\textbf{Claude Opus~5}} & \multicolumn{2}{c}{\textbf{Claude Opus~5.5}} & \multicolumn{2}{c}{\textbf{GPT-6 Astra}} \\
\cmidrule(lr){2-3}\cmidrule(lr){4-5}\cmidrule(lr){6-7}
 & w/o evo & w/ evo & w/o evo & w/ evo & w/o evo & w/ evo \\
\midrule
\multicolumn{7}{l}{\emph{Position perturbation}} \\
0~~Open middle drawer & 90\% & 90\% & 100\% & 100\% & 96\% & 96\% \\
1~~Bowl $\to$ stove & 94\% & 94\% & 90\% & \textbf{100\%} & 92\% & 92\% \\
2~~Wine bottle $\to$ drawer top & 98\% & 98\% & 100\% & 100\% & 100\% & 100\% \\
3~~Open drawer, bowl inside & 100\% & 100\% & 100\% & 100\% & 100\% & 100\% \\
4~~Bowl $\to$ drawer top & 100\% & 100\% & 100\% & 100\% & 92\% & 92\% \\
5~~Push plate to stove & 88\% & 88\% & 84\% & 84\% & 98\% & 98\% \\
6~~Cream cheese $\to$ bowl & 100\% & 100\% & 100\% & 100\% & 92\% & 92\% \\
7~~Turn on stove & 100\% & 100\% & 100\% & 100\% & 100\% & 100\% \\
8~~Bowl $\to$ plate & 100\% & 100\% & 96\% & 96\% & 100\% & 100\% \\
9~~Wine bottle $\to$ rack & 100\% & 100\% & 100\% & 100\% & 100\% & 100\% \\
\cmidrule(lr){2-7}
\emph{Average} & 97.0\% & 97.0\% & 97.0\% & 98.0\% & 97.0\% & 97.0\% \\
\midrule
\multicolumn{7}{l}{\emph{Task perturbation}} \\
0~~Open bottom drawer & 0\% & 0\% & 100\% & 100\% & 82\% & 82\% \\
1~~Plate $\to$ stove & 100\% & 100\% & 100\% & 100\% & 100\% & 100\% \\
2~~Wine bottle $\to$ bowl & 96\% & 96\% & 100\% & 100\% & 100\% & 100\% \\
3~~Open drawer, cheese inside & 98\% & 98\% & 100\% & 100\% & 100\% & 100\% \\
4~~Plate $\to$ drawer top & 88\% & 88\% & 100\% & 100\% & 100\% & 100\% \\
5~~Push cheese to stove & 98\% & 98\% & 100\% & 100\% & 98\% & 98\% \\
6~~Wine bottle $\to$ bowl & 94\% & 94\% & 100\% & 100\% & 100\% & 100\% \\
7~~Turn off stove & 100\% & 100\% & 94\% & 94\% & 92\% & 92\% \\
8~~Wine bottle $\to$ plate & 94\% & 94\% & 100\% & 100\% & 100\% & 100\% \\
9~~Cream cheese $\to$ rack & 100\% & 100\% & 100\% & 100\% & 96\% & 96\% \\
\cmidrule(lr){2-7}
\emph{Average} & 86.8\% & 86.8\% & 99.4\% & 99.4\% & 96.8\% & 96.8\% \\
\bottomrule
\end{tabular}
\end{table}

\begin{table}[H]
\centering
\footnotesize
\caption{\textbf{Per-task model ablation on LIBERO-Pro Spatial.} As in Table~\ref{tab:libero-per-task-object}, for LIBERO-Pro Spatial. Spatial tasks pick the black bowl at the listed location and place it on the plate.}
\label{tab:libero-per-task-spatial}
\setlength{\tabcolsep}{5pt}
\begin{tabular}{lcccccc}
\toprule
\textbf{Task} & \multicolumn{2}{c}{\textbf{Claude Opus~5}} & \multicolumn{2}{c}{\textbf{Claude Opus~5.5}} & \multicolumn{2}{c}{\textbf{GPT-6 Astra}} \\
\cmidrule(lr){2-3}\cmidrule(lr){4-5}\cmidrule(lr){6-7}
 & w/o evo & w/ evo & w/o evo & w/ evo & w/o evo & w/ evo \\
\midrule
\multicolumn{7}{l}{\emph{Position perturbation}} \\
0~~Between plate \& ramekin & 98\% & 98\% & 100\% & 100\% & 98\% & 98\% \\
1~~Next to ramekin & 96\% & 96\% & 100\% & 100\% & 100\% & 100\% \\
2~~Table center & 100\% & 100\% & 100\% & 100\% & 96\% & 96\% \\
3~~On cookie box & 98\% & 98\% & 100\% & 100\% & 96\% & 96\% \\
4~~In cabinet top drawer & 96\% & 96\% & 100\% & 100\% & 82\% & \textbf{98\%} \\
5~~On ramekin & 90\% & 90\% & 100\% & 100\% & 100\% & 100\% \\
6~~Next to cookie box & 100\% & 100\% & 100\% & 100\% & 100\% & 100\% \\
7~~On stove & 100\% & 100\% & 100\% & 100\% & 98\% & 98\% \\
8~~Next to plate & 100\% & 100\% & 94\% & 94\% & 100\% & 100\% \\
9~~On wooden cabinet & 100\% & 100\% & 100\% & 100\% & 100\% & 100\% \\
\cmidrule(lr){2-7}
\emph{Average} & 97.8\% & 97.8\% & 99.4\% & 99.4\% & 97.0\% & 98.6\% \\
\midrule
\multicolumn{7}{l}{\emph{Task perturbation}} \\
0~~Not btw. plate \& ramekin & 98\% & 98\% & 100\% & 100\% & 92\% & 92\% \\
1~~Next to cookie box & 92\% & 92\% & 100\% & 100\% & 100\% & 100\% \\
2~~Next to plate & 90\% & 90\% & 92\% & 92\% & 92\% & 92\% \\
3~~On cabinet top & 100\% & 100\% & 100\% & 100\% & 100\% & 100\% \\
4~~On wooden cabinet top & 100\% & 100\% & 100\% & 100\% & 100\% & 100\% \\
5~~On cookie box & 100\% & 100\% & 100\% & 100\% & 100\% & 100\% \\
6~~On stove & 100\% & 100\% & 100\% & 100\% & 100\% & 100\% \\
7~~On cabinet top & 100\% & 100\% & 100\% & 100\% & 100\% & 100\% \\
8~~Next to ramekin & 100\% & 100\% & 100\% & 100\% & 96\% & 96\% \\
9~~On stove & 100\% & 100\% & 100\% & 100\% & 96\% & 96\% \\
\cmidrule(lr){2-7}
\emph{Average} & 98.0\% & 98.0\% & 99.2\% & 99.2\% & 97.6\% & 97.6\% \\
\bottomrule
\end{tabular}
\end{table}

\begin{table}[H]
\centering
\footnotesize
\caption{\textbf{Per-task model ablation on LIBERO-Pro Long.} As in Table~\ref{tab:libero-per-task-object}, for LIBERO-Pro Long.}
\label{tab:libero-per-task-long}
\setlength{\tabcolsep}{5pt}
\begin{tabular}{lcccccc}
\toprule
\textbf{Task}
& \multicolumn{2}{c}{\textbf{Claude Opus~5}}
& \multicolumn{2}{c}{\textbf{Claude Opus~5.5}}
& \multicolumn{2}{c}{\textbf{GPT-6 Astra}} \\
\cmidrule(lr){2-3}
\cmidrule(lr){4-5}
\cmidrule(lr){6-7}
& w/o evo & w/ evo
& w/o evo & w/ evo
& w/o evo & w/ evo \\
\midrule

\multicolumn{7}{l}{\emph{Position perturbation}} \\
0~~Soup + sauce $\to$ basket
  & 94\% & 94\% & 98\% & 98\% & 64\% & \textbf{76\%} \\
1~~Cheese + butter $\to$ basket
  & 98\% & 98\% & 100\% & 100\% & 94\% & 94\% \\
2~~Stove on + moka pot
  & 98\% & 98\% & 98\% & 98\% & 96\% & 96\% \\
3~~Bowl $\to$ drawer, close
  & 52\% & \textbf{78\%} & 98\% & 98\% & 90\% & 90\% \\
4~~Mugs $\to$ L/R plates
  & 98\% & 98\% & 100\% & 100\% & 90\% & 90\% \\
5~~Book $\to$ caddy back
  & 96\% & 96\% & 84\% & 84\% & 96\% & 96\% \\
6~~Mug $\to$ plate, pudding right
  & 98\% & 98\% & 98\% & 98\% & 84\% & 84\% \\
7~~Soup + cheese $\to$ basket
  & 88\% & \textbf{92\%} & 98\% & 98\% & 82\% & \textbf{84\%} \\
8~~Both moka pots $\to$ stove
  & 88\% & \textbf{92\%} & 100\% & 100\% & 82\% & 82\% \\
9~~Mug $\to$ microwave, close
  & 0\% & \textbf{26\%} & 96\% & 96\% & 30\% & \textbf{54\%} \\
\cmidrule(lr){2-7}
\emph{Average}
  & 81.0\% & 87.0\% & 97.0\% & 97.0\% & 80.8\% & 84.6\% \\

\midrule
\multicolumn{7}{l}{\emph{Task perturbation}} \\
0~~Cheese + sauce $\to$ basket
  & 98\% & 98\% & 100\% & 100\% & 94\% & 94\% \\
1~~Soup + butter $\to$ basket
  & 98\% & 98\% & 88\% & 88\% & 92\% & 92\% \\
2~~Stove on + pan
  & 94\% & 94\% & 100\% & 100\% & 100\% & 100\% \\
3~~Bottle $\to$ drawer, close
  & 72\% & \textbf{100\%} & 100\% & 100\% & 96\% & 96\% \\
4~~Mugs $\to$ L/R plates, swapped
  & 100\% & 100\% & 100\% & 100\% & 100\% & 100\% \\
5~~Cup $\to$ caddy back
  & 86\% & 86\% & 100\% & 100\% & 100\% & 100\% \\
6~~Red mug $\to$ plate, pudding
  & 86\% & 86\% & 100\% & 100\% & 100\% & 100\% \\
7~~Ketchup + cheese $\to$ basket
  & 100\% & 100\% & 100\% & 100\% & 100\% & 100\% \\
8~~Left moka pot $\to$ stove
  & 100\% & 100\% & 100\% & 100\% & 100\% & 100\% \\
9~~Mug $\to$ microwave, close
  & 0\% & \textbf{42\%} & 100\% & 100\% & 68\% & 68\% \\
\cmidrule(lr){2-7}
\emph{Average}
  & 83.4\% & 90.4\% & 98.8\% & 98.8\% & 95.0\% & 95.0\% \\

\bottomrule
\end{tabular}
\end{table}

\subsection{Model Ablation on Robosuite}
\label{sec:appendix:model-ablation-robosuite}

\begin{table}[H]
\centering
\footnotesize
\caption{\textbf{Model ablation on Robosuite.} Success on all seven Robosuite tasks over 100 held-out trials per task. Table~\ref{tab:ablation-backbone-robosuite} shows the three hardest tasks.}
\label{tab:model-ablation-robosuite-full}
\setlength{\tabcolsep}{7pt}
\begin{tabular}{lcccc}
\toprule
\textbf{Task} & \textbf{Claude Opus~4.6} & \textbf{Claude Opus~5} & \textbf{Claude Opus~5.5} & \textbf{GPT-6 Astra} \\
\midrule
cube\_lift         & 97\%  & 100\% & 100\% & 100\% \\
cube\_stack        & 99\%  & 100\% & 100\% & 99\%  \\
cube\_restack      & 100\% & 100\% & 100\% & 100\% \\
spill\_wipe        & 99\%  & 99\%  & 100\% & 99\%  \\
two\_arm\_handover & 92\%  & 99\%  & 100\% & 95\%  \\
two\_arm\_lift     & 71\%  & 96\%  & 100\% & 100\% \\
nut\_assembly      & 9\%   & 98\%  & 100\% & 99\%  \\
\midrule
\textit{Mean}      & \textit{81\%} & \textit{99\%} & \textit{100\%} & \textit{99\%} \\
\bottomrule
\end{tabular}
\end{table}

\subsection{Model Ablation on LIBERO-Pro Goal under Position Perturbation}
\label{sec:appendix:goal-swap-iterative-debug}
\label{sec:appendix:goal-swap-fresh-run}

\begin{table}[H]
\centering
\footnotesize
\caption{\textbf{Success and cost of all models in Figure~\ref{fig:model-ablation}.} Held-out success on the 10 LIBERO-Pro Goal tasks under the position perturbation (50 seeds each), using \name\ iterative debugging without evolutionary search and with an initially empty skill library. Cost is estimated at list API prices; agent time is the mean time per task.}
\label{tab:model-cost-summary}
\setlength{\tabcolsep}{6pt}
\begin{tabular}{llcccc}
\toprule
\textbf{Model} & \textbf{Open weights} & \textbf{Success} & \textbf{Cost / task} & \textbf{Total cost} & \textbf{Agent time / task (min)} \\
\midrule
Claude Opus~5.5      &            & 99.2\% & \$18.29 & \$183 & 51 \\
Claude Sonnet~5.5    &            & 98.8\% & \$12.11 & \$121 & 56 \\
Claude Fable~5.1     &            & 96.6\% & \$42.24 & \$422 & 68 \\
Claude Opus~5        &            & 90.6\% & \$59.72 & \$597 & 76 \\
Claude Haiku~5.5     &            & 68.0\% & \$3.97 & \$40 & 54 \\
Claude Opus~4.6      &            & 55.6\% & \$23.58 & \$236 & 61 \\
\midrule
GPT-6 Astra          &            & 94.4\% & \$35.37 & \$354 & 31 \\
GPT-6.1 Sol          &            & 88.0\% & \$9.52 & \$95 & 49 \\
GPT-6 Sol            &            & 59.2\% & \$7.64 & \$76 & 42 \\
GPT-6 Luna           &            & 14.8\% & \$0.83 & \$8 & 47 \\
\midrule
Kimi~K3              & \checkmark & 78.4\% & \$58.19 & \$582 & 139 \\
Qwen~3.8 Max         & \checkmark & 56.0\% & \$39.82 & \$398 & 103 \\
\bottomrule
\end{tabular}
\end{table}

\begin{table}[H]
\centering
\scriptsize
\caption{\textbf{Per-task model ablation on LIBERO-Pro Goal, position perturbation.} Iterative debugging results without evolutionary search, as in Figure~\ref{fig:model-ablation}. Stage~1 iterative debugging on seeds 51--65; Stage~2 held-out evaluation on seeds 1--50 is reported below. Claude Opus~4.6 (max effort), Claude Opus~5 and Claude Opus~5.5 run through Claude Code with a 1M-token context; GPT-6 Astra runs through Codex with a 258k-token context. The skill library is initially empty.}
\label{tab:model_change_fresh}
\setlength{\tabcolsep}{3pt}
\begin{tabular}{>{\scriptsize}lcccc}
\toprule
\textbf{Task} & \textbf{Claude Opus~4.6} & \textbf{Claude Opus~5} & \textbf{Claude Opus~5.5} & \textbf{GPT-6 Astra} \\
\midrule
Open middle drawer           & 0\%   & \textbf{100\%} & \textbf{100\%} & 98\% \\
Bowl on stove                & 0\%   & \textbf{100\%} & \textbf{100\%} & 92\% \\
Wine bottle on cabinet       & 94\%  & \textbf{100\%} & \textbf{100\%} & \textbf{100\%} \\
Bowl inside top drawer       & 0\%   & \textbf{100\%} & \textbf{100\%} & \textbf{100\%} \\
Bowl on cabinet top          & 96\%  & \textbf{100\%} & \textbf{100\%} & \textbf{100\%} \\
Push plate to stove          & 0\%   & 14\% & \textbf{92\%} & 54\% \\
Cream cheese in bowl         & 82\%  & \textbf{100\%} & \textbf{100\%} & \textbf{100\%} \\
Turn on stove                & \textbf{100\%} & \textbf{100\%} & \textbf{100\%} & \textbf{100\%} \\
Bowl on plate                & \textbf{100\%} & 94\% & \textbf{100\%} & \textbf{100\%} \\
Wine bottle on rack          & 84\%  & 98\% & \textbf{100\%} & \textbf{100\%} \\
\midrule
\textit{Mean}                & \textit{55.6\%} & \textit{90.6\%} & \textit{99.2\%} & \textit{94.4\%} \\
\bottomrule
\end{tabular}
\end{table}

\subsection{Model Ablation Token Usage on LIBERO-Pro Object, Goal, and Spatial}
\label{sec:appendix:libero-pro-run-tokens}
\begin{table}[H]
\centering
\scriptsize
\caption{\textbf{Token usage and cost on LIBERO-Pro Object, Goal, and Spatial.} Token usage, estimated API cost, and compute for the 60-task
LIBERO-Pro runs. Total tokens include all input and output. Evolutionary search was triggered on tasks below 90\% success.
GPT-6 Astra (xhigh, Codex) is the Object, Goal, and Spatial phase of a run that continues on LIBERO-Pro Long;
its coordinator is split at the phase boundary, and Codex's approval-reviewer calls are excluded.
Claude Opus~4.6 (max effort) is likewise the first phase of a run that continues on LIBERO-Pro Long; its coordinator is allocated by
agent cost, and its token counts are exact, including Claude Code background calls. Claude Opus~4.6 costs are from a re-run with the current pipeline; its success rates are reported from the original run (Tables~\ref{tab:libero-pro-ablation-pos} and~\ref{tab:libero-pro-ablation-task}).}
\label{tab:libero-pro-60-usage}
\setlength{\tabcolsep}{3pt}
\begin{tabular}{>{\scriptsize}lcccc}
\toprule
\textbf{Metric} & \textbf{Claude Opus~4.6} & \textbf{Claude Opus~5} & \textbf{Claude Opus~5.5} & \textbf{GPT-6 Astra} \\
\midrule
\multicolumn{5}{l}{\emph{Without evolutionary search}} \\
Total tokens                 & 2,009.3M & 2,542.9M & 1,538.3M & 522.1M \\
Cache-read tokens            & 1,966.5M & 2,471.7M & 1,493.5M & 505.4M \\
Output tokens                & 8.18M    & 13.44M   & 8.02M    & 2.57M \\
Est.\ cost (USD)             & \$1,376  & \$1,925  & \$650    & \$776 \\
Cost per task (USD)          & \$22.9   & \$32.1   & \$10.8   & \$12.9 \\
\midrule
Sim rollouts                 & 12,214   & 10,941   & 10,555   & 7,905 \\
LLM calls                    & 9,405    & 8,535    & 7,896    & 5,950 \\
Actor time per task (min)    & 79       & 60       & 46       & 27 \\
\midrule
\multicolumn{5}{l}{\emph{Evolutionary search only, additional usage}} \\
Tasks searched               & 31       & 3        & 7        & 2 \\
Total tokens                 & 4,937.7M & 54.7M    & 246.7M   & 17.4M \\
Cache-read tokens            & 4,908.6M & 52.7M    & 227.0M   & 17.1M \\
Output tokens                & 5.17M    & 0.24M    & 1.12M    & 0.06M \\
Est.\ cost (USD)             & \$2,761  & \$43     & \$160    & \$23 \\
Cost per searched task (USD) & \$89.1   & \$14.4   & \$22.9   & \$11.4 \\
\midrule
Sim rollouts                 & 9,529    & 454      & 1,359    & 344 \\
LLM calls                    & 17,037   & 352      & 746      & 236 \\
Actor time per task (min)    & 147      & 43       & 32       & 39 \\
\bottomrule
\end{tabular}
\end{table}

\begin{table}[H]
\centering
\scriptsize
\caption{\textbf{Per-suite success and cost on LIBERO-Pro Object, Goal, and Spatial.} Held-out success (SR), list-price cost per task, and actor time per task for each suite and perturbation in the 60-task LIBERO-Pro runs, without evolutionary search. Actor, verifier, and rerun usage is attributed to its task; the coordinator is allocated to tasks in proportion to their agent cost.}
\label{tab:libero-pro-per-suite-usage}
\setlength{\tabcolsep}{3pt}
\resizebox{\textwidth}{!}{%
\begin{tabular}{lcccccccccccc}
\toprule
\textbf{Suite · perturbation} & \multicolumn{3}{c}{\textbf{Claude Opus~4.6}} & \multicolumn{3}{c}{\textbf{Claude Opus~5}} & \multicolumn{3}{c}{\textbf{Claude Opus~5.5}} & \multicolumn{3}{c}{\textbf{GPT-6 Astra}} \\
\cmidrule(lr){2-4} \cmidrule(lr){5-7} \cmidrule(lr){8-10} \cmidrule(lr){11-13}
 & SR & \$/task & min/task & SR & \$/task & min/task & SR & \$/task & min/task & SR & \$/task & min/task \\
\midrule
Goal · Position & -- & \$35.0 & 112 & 97.0\% & \$35.1 & 76 & 97.0\% & \$13.3 & 56 & 97.0\% & \$13.7 & 27 \\
Goal · Task & -- & \$28.1 & 100 & 86.8\% & \$39.3 & 78 & 99.4\% & \$11.5 & 46 & 96.8\% & \$12.0 & 25 \\
Object · Position & -- & \$20.1 & 68 & 98.6\% & \$29.2 & 48 & 98.2\% & \$9.2 & 41 & 99.0\% & \$14.8 & 27 \\
Object · Task & -- & \$11.0 & 36 & 100.0\% & \$24.7 & 46 & 100.0\% & \$9.4 & 40 & 99.4\% & \$13.3 & 28 \\
Spatial · Position & -- & \$24.7 & 90 & 97.8\% & \$37.1 & 57 & 99.4\% & \$11.5 & 45 & 97.0\% & \$13.0 & 29 \\
Spatial · Task & -- & \$18.8 & 70 & 98.0\% & \$27.2 & 52 & 99.2\% & \$10.1 & 45 & 97.6\% & \$10.8 & 25 \\
\midrule
\emph{Position (avg)} & -- & \$26.6 & 90 & 97.8\% & \$33.8 & 60 & 98.2\% & \$11.3 & 47 & 97.7\% & \$13.8 & 28 \\
\emph{Task (avg)} & -- & \$19.3 & 69 & 94.9\% & \$30.4 & 59 & 99.5\% & \$10.4 & 44 & 97.9\% & \$12.0 & 26 \\
\midrule
\emph{Goal (avg)} & -- & \$31.6 & 106 & 91.9\% & \$37.2 & 77 & 98.2\% & \$12.4 & 51 & 96.9\% & \$12.8 & 26 \\
\emph{Object (avg)} & -- & \$15.5 & 52 & 99.3\% & \$26.9 & 47 & 99.1\% & \$9.3 & 41 & 99.2\% & \$14.1 & 28 \\
\emph{Spatial (avg)} & -- & \$21.7 & 80 & 97.9\% & \$32.1 & 55 & 99.3\% & \$10.8 & 45 & 97.3\% & \$11.9 & 27 \\
\midrule
\emph{All 60} & -- & \$22.9 & 79 & 96.4\% & \$32.1 & 60 & 98.9\% & \$10.8 & 46 & 97.8\% & \$12.9 & 27 \\
\bottomrule
\end{tabular}}
\end{table}

\subsection{Model Ablation Token Usage on LIBERO-Pro Long}
\label{sec:appendix:libero-10-run-tokens}
\begin{table}[H]
\centering
\scriptsize
\caption{\textbf{Token usage and cost on LIBERO-Pro Long.} As in Table~\ref{tab:libero-pro-60-usage}, for the 20-task LIBERO-Pro Long runs. Token usage is transcript-derived, and total tokens include input and estimated output; Claude Opus~4.6 token counts are exact. All four runs used the skill library from their LIBERO-Pro phase.}
\label{tab:libero-10-usage}
\setlength{\tabcolsep}{3pt}
\begin{tabular}{>{\scriptsize}lcccc}
\toprule
\textbf{Metric} & \textbf{Claude Opus~4.6} & \textbf{Claude Opus~5} & \textbf{Claude Opus~5.5} & \textbf{GPT-6 Astra} \\
\midrule
\multicolumn{5}{l}{\emph{Without evolutionary search}} \\
Total tokens                 & 550.5M   & 1,544.2M & 800.2M & 490.5M \\
Cache-read tokens            & 539.1M   & 1,487.6M & 774.6M & 479.3M \\
Output tokens                & 1.98M    & 5.28M    & 3.56M  & 2.03M \\
Est.\ cost (USD)             & \$411    & \$1,206  & \$340  & \$672 \\
Cost per task (USD)          & \$20.6   & \$60.3   & \$17.0 & \$33.6 \\
\midrule
Sim rollouts                 & 3,169    & 4,722    & 3,870  & 3,826 \\
LLM calls                    & 2,207    & 5,451    & 3,403  & 4,413 \\
Actor time per task (min)    & 73       & 137      & 70     & 73 \\
\midrule
\multicolumn{5}{l}{\emph{Evolutionary search only, additional usage}} \\
Tasks searched               & 19       & 8        & 2      & 6 \\
Total tokens                 & 5,119.9M & 1,111.1M & 72.5M  & 124.4M \\
Cache-read tokens            & 5,090.3M & 1,070.3M & 66.7M  & 122.9M \\
Output tokens                & 4.68M    & 1.78M    & 0.33M  & 0.20M \\
Est.\ cost (USD)             & \$2,792  & \$826    & \$47   & \$146 \\
Cost per searched task (USD) & \$146.9  & \$103.2  & \$23.6 & \$24.4 \\
\midrule
Sim rollouts                 & 7,052    & 2,570    & 415    & 1,274 \\
LLM calls                    & 16,949   & 2,427    & 214    & 1,163 \\
Actor time per searched task (min) & 256 & 291 & 114 & 143 \\
\bottomrule
\end{tabular}
\end{table}

\begin{table}[H]
\centering
\scriptsize
\caption{\textbf{Per-perturbation success and cost on LIBERO-Pro Long.} As in Table~\ref{tab:libero-pro-per-suite-usage}, for the 20-task LIBERO-Pro Long runs. All four runs used the skill library from their LIBERO-Pro phase.}
\label{tab:libero-10-per-perturbation-usage}
\setlength{\tabcolsep}{3pt}
\resizebox{\textwidth}{!}{%
\begin{tabular}{lcccccccccccc}
\toprule
\textbf{Perturbation} & \multicolumn{3}{c}{\textbf{Claude Opus~4.6}} & \multicolumn{3}{c}{\textbf{Claude Opus~5}} & \multicolumn{3}{c}{\textbf{Claude Opus~5.5}} & \multicolumn{3}{c}{\textbf{GPT-6 Astra}} \\
\cmidrule(lr){2-4} \cmidrule(lr){5-7} \cmidrule(lr){8-10} \cmidrule(lr){11-13}
 & SR & \$/task & min/task & SR & \$/task & min/task & SR & \$/task & min/task & SR & \$/task & min/task \\
\midrule
Position & -- & \$20.3 & 73 & 81.0\% & \$73.7 & 154 & 97.0\% & \$17.7 & 74 & 80.8\% & \$37.4 & 84 \\
Task & -- & \$20.8 & 72 & 83.4\% & \$46.9 & 119 & 98.8\% & \$16.3 & 66 & 95.0\% & \$29.8 & 63 \\
\midrule
\emph{All 20} & -- & \$20.6 & 73 & 82.2\% & \$60.3 & 137 & 97.9\% & \$17.0 & 70 & 87.9\% & \$33.6 & 73 \\
\bottomrule
\end{tabular}}
\end{table}

\subsection{LIBERO-Pro Long Iterative Debugging}
\begin{table}[H]
\centering
\scriptsize
\caption{\textbf{Iterative debugging on LIBERO-Pro Long.} Stage~2 held-out success on seeds 1--50 under the position (Pos) and task (Task) perturbations, without evolutionary search, using Claude Opus~4.6 in Claude Code. \textbf{ZS}: zero-shot result; \textbf{ID}: result after iterative debugging.}
\label{tab:best_of}
\setlength{\tabcolsep}{3pt}
\begin{tabular}{>{\scriptsize}lcccc}
\toprule
& \multicolumn{2}{c}{\textbf{Pos}} & \multicolumn{2}{c}{\textbf{Task}} \\
\cmidrule(lr){2-3}\cmidrule(lr){4-5}
\textbf{Task} & \textbf{ZS} & \textbf{ID} & \textbf{ZS} & \textbf{ID} \\
\midrule
Soup + sauce to basket (S2)      & 0\% & \textbf{22\%} & 70\% & \textbf{98\%} \\
Cheese + butter to basket (S2)   & 94\% & 94\% & 86\% & 86\% \\
Stove knob + moka pot (S3)       & 100\% & 100\% & 26\% & \textbf{68\%} \\
Bowl/bottle to drawer (S4)       & 0\% & 0\% & 0\% & 0\% \\
Mugs to plates L/R (S5)          & 32\% & 32\% & 0\% & 0\% \\
Book/cup to caddy back (Study1)  & 0\% & 0\% & 26\% & 26\% \\
Mug to plate + pudding (S6)      & 0\% & \textbf{36\%} & 4\% & \textbf{34\%} \\
Soup/ketchup + cheese (S1)       & 0\% & \textbf{94\%} & 70\% & \textbf{94\%} \\
Moka pot(s) to stove (S8)        & 0\% & 0\% & 100\% & 100\% \\
Mug to microwave + close (S6)    & 0\% & \textbf{54\%} & 0\% & \textbf{14\%} \\
\midrule
\textit{Mean}                    & \textit{22.6\%} & \textbf{\textit{43.2\%}} & \textit{38.3\%} & \textbf{\textit{52.0\%}} \\
\bottomrule
\end{tabular}
\end{table}

\subsection{Iterative Debugging Variance on LIBERO-Pro Goal under Position Perturbation}
\begin{table}[H]
\centering
\scriptsize
\caption{\textbf{Iterative debugging variance on LIBERO-Pro Goal, position perturbation.} Stage~2 held-out success on seeds 1--50 for three independent iterative debugging runs with Claude Opus~4.6 in Claude Code. Each run starts from a clean skill library with no prior knowledge carried over.}
\label{tab:goal_swap_variance}
\setlength{\tabcolsep}{3pt}
\begin{tabular}{>{\scriptsize}lccccccc}
\toprule
& \multicolumn{3}{c}{\textbf{Held-Out Success Rate}} & & & & \\
\cmidrule(lr){2-4}
\textbf{Task} & \textbf{Run 1} & \textbf{Run 3} & \textbf{Run 4} & \textbf{Mean} & \textbf{Std} & \textbf{Min} & \textbf{Max} \\
\midrule
Open middle drawer           & 0\% & \textbf{100\%} & 0\% & 33.3\% & 47.1\% & 0\% & 100\% \\
Bowl on stove                & 8\% & 0\% & 0\% & 2.7\% & 3.8\% & 0\% & 8\% \\
Wine bottle on cabinet       & \textbf{100\%} & 96\% & 96\% & 97.3\% & 1.9\% & 96\% & 100\% \\
Open drawer + bowl inside    & \textbf{62\%} & 0\% & 0\% & 20.7\% & 29.2\% & 0\% & 62\% \\
Bowl on cabinet top          & 96\% & 80\% & 96\% & 90.7\% & 7.5\% & 80\% & 96\% \\
Push plate to stove          & 0\% & 0\% & 0\% & 0.0\% & 0.0\% & 0\% & 0\% \\
Cream cheese in bowl         & \textbf{100\%} & \textbf{100\%} & \textbf{100\%} & 100.0\% & 0.0\% & 100\% & 100\% \\
Turn on stove                & \textbf{100\%} & \textbf{100\%} & \textbf{100\%} & 100.0\% & 0.0\% & 100\% & 100\% \\
Bowl on plate                & 92\% & \textbf{100\%} & \textbf{100\%} & 97.3\% & 3.8\% & 92\% & 100\% \\
Wine bottle on rack          & 70\% & 98\% & \textbf{100\%} & 89.3\% & 13.7\% & 70\% & 100\% \\
\midrule
\textit{Mean}                & \textit{62.8\%} & \textit{67.4\%} & \textit{59.2\%} & \textbf{\textit{63.1\%}} & \textit{3.4\%} & \textit{59.2\%} & \textit{67.4\%} \\
\bottomrule
\end{tabular}
\end{table}

\clearpage

\section{Case Study Details}
\label{sec:appendix:case-study}

\begin{table}[H]
\centering
\footnotesize
\caption{\textbf{Case-study success rates.} Success over 50 held-out seeds for each model on the case-study tasks, from the case-study runs (separate from the per-task tables in Appendix~\ref{sec:appendix:ablations}). L90 denotes LIBERO-90; Goal, Spatial, and Long denote LIBERO-Pro suites, with Pos and Task denoting the position and task perturbations. Opus refers to Claude Opus, Astra refers to GPT-6 Astra. Evo refers to evolutionary search, run on tasks below 90\% success rate.}
\label{tab:case-study-sr}
\setlength{\tabcolsep}{3pt}
\begin{tabular}{llcccccc}
\toprule
\textbf{Task} & \textbf{Suite} & \textbf{\shortstack{Opus\\4.6}} & \textbf{\shortstack{Opus\\5}} & \textbf{\shortstack{Opus\\5.5}} & \textbf{\shortstack{Opus 5.5\\+Evo}} & \textbf{Astra} & \textbf{\shortstack{Astra\\+Evo}} \\
\midrule
Book under cabinet shelf  & L90 \#89          & 0\%  & 2\%  & 68\%  & 100\% & 22\%  & 96\%  \\
Open bottom drawer        & L90 \#6           & 0\%  & 0\%  & 90\%  & 90\%  & 92\%  & 100\% \\
Push plate to stove front & Goal \#5, Pos     & 0\%  & 0\%  & 84\%  & 94\%  & 62\%  & 92\%  \\
Bowl on stove             & Goal \#1, Pos     & 56\% & 64\% & 90\%  & 90\%  & 90\%  & 90\%  \\
Mug in microwave, close   & Long \#9, Pos     & 54\% & 16\% & 96\%  & 96\%  & 98\%  & 98\%  \\
Plate on stove            & Goal \#1, Task    & 0\%  & 54\% & 100\% & 100\% & 100\% & 100\% \\
Akita bowl on plate       & Spatial \#2, Task & 46\% & 62\% & 92\%  & 92\%  & 92\%  & 92\%  \\
\bottomrule
\end{tabular}
\end{table}

\begin{figure}[H]
\centering
\includegraphics[width=\textwidth]{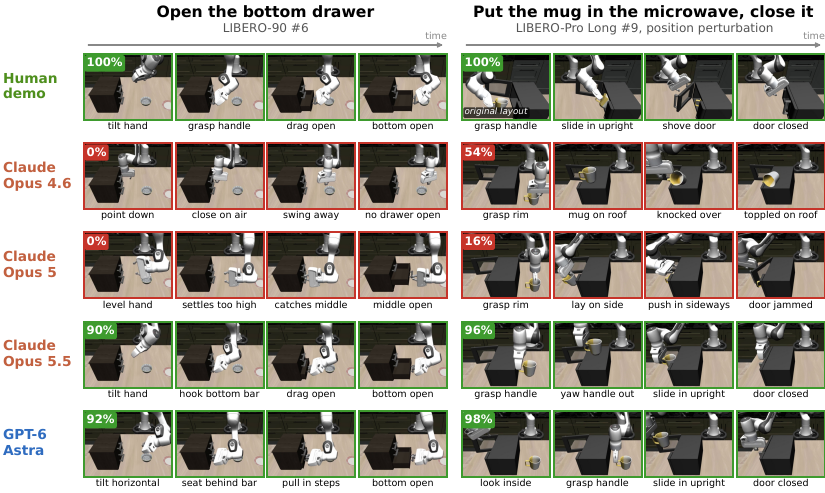}
\caption{\textbf{Case study: two more tasks.} Same layout as Figure~\ref{fig:case-study}. The mug demonstration is from the original LIBERO-Long layout, with the microwave on the other side.}
\label{fig:case-study-full}
\end{figure}

\paragraph{Open the bottom drawer.} The bottom handle is too low for a level hand. The human tilts the hand and pulls the handle. Opus~5.5 and Astra do the same: they close the gripper, tilt it, hook the fingertips behind the handle, and drag the drawer open (90\% and 92\%). Opus~5 keeps the hand level, catches the middle handle, and opens the middle drawer (0\%). Opus~4.6 never reaches a handle (0\%).

\paragraph{Mug in the microwave.} Opus~5.5 and Astra grasp the handle and slide the mug in upright, as the human does (96\% and 98\%); Astra first looks into the microwave with the wrist camera. Opus~5 grasps the rim, so it must lay the mug on its side to fit it through the opening; the handle then sticks out and the door jams (16\%). The Opus~4.6 episode treats the microwave as top-loading and leaves the mug on its roof.

\section{Agent Pipeline Skills}
\label{sec:appendix:prompts-skills}

\subsection{Codebase Overview and Skill File Locations}
\label{sec:appendix:skill-library-prompts}

The code is publicly available at \url{https://github.com/NVlabs/ASPIRE}. \\

\dirtree{%
.1 ASPIRE/.
.2 src/cap\_harness/.
.3 api.py, contracts.py \DTcomment{public CaP program API}.
.3 run.py, batch.py \DTcomment{single-run and batched runners}.
.3 providers/ \DTcomment{sam3, graspnet, curobo, pyroki}.
.3 libero/ robosuite/ behavior/ yam\_real/ \DTcomment{benchmark adapters}.
.3 validation/ \DTcomment{held-out evaluators}.
.2 docs/ \DTcomment{api-reference.md, run-artifacts.md, setup}.
.2 configs/ \DTcomment{environment pins, validation profiles}.
.2 scripts/ \DTcomment{bootstrap\_*.sh, supervise\_services.sh}.
.2 .claude/skills/.
.3 iterative-debugging/ \DTcomment{fix loop + skill library}.
.3 evosearch/ \DTcomment{evolutionary search}.
.2 .codex/skills/ \DTcomment{Codex entry points}.
.2 .agents/skills/ \DTcomment{System 2 skills}.
.2 AGENTS.md \DTcomment{System 2 execution rules}.
}

\paragraph{Pipeline files.}
Each pipeline skill is a directory with the same four prompt files, entered through
\texttt{SKILL.md}:
\texttt{main-agent-prompt.md} (coordinator), \texttt{subagent-prompt.md} (actor agent: the
forbidden-API rules, findings schema, and debugging protocol),
\texttt{verifier-prompt.md} (iterative-debugging only), and \texttt{clean-task-slate.md}.
The orchestration code is in \texttt{scripts/} (\texttt{dispatch.py} fills the templates per task,
\texttt{run\_validation.py} runs the held-out partition, \texttt{lint\_skills.py} checks a proposal
before a verifier is spent on it, \texttt{campaign.py} holds campaign settings).
BEHAVIOR-1K uses a different protocol, in \texttt{iterative-debugging/behavior/}.

\paragraph{Skill library.}
The skill library is stored in
\texttt{.claude/skills/iterative-debugging/skills/}: \texttt{grasp.md}, \texttt{localize.md},
\texttt{transport.md}, \texttt{manipulation.md}, and \texttt{task-exploration.md}, indexed by
\texttt{skills/README.md}.
Actor agents report candidate repairs in a structured findings file (observed failure mode, validated repair, transferable patterns, task-specific quirks, and success rate on debug configurations). A separate verifier agent must reproduce the fix from the proposal alone, and only then does the coordinator promote it. Entries carry provenance (source task, development seeds, held-out result).

\newpage

\end{document}